\documentclass[letterpaper,11pt,onecolumn,conference,romanappendices]{ieeeconf2}
\IEEEoverridecommandlockouts
\usepackage{amsmath,amssymb,theorem}
\usepackage{graphicx}
\graphicspath{ {./images/} }
\usepackage{algorithm}
\usepackage{algpseudocode}
\usepackage{psfrag}
\usepackage{cite}
\usepackage[margin=1in]{geometry}
\usepackage{tabularx}
\usepackage{booktabs}
\usepackage{titlesec}
\usepackage{hyperref}

\usepackage{color,xspace}
\usepackage{subfigure}
\usepackage{microtype}
\usepackage[table]{xcolor}
\newtheorem{theorem}{Theorem}[section]

\newtheorem{remark}[theorem]{Remark}
\newtheorem{example}[theorem]{Example}

\newtheorem{proposition}[theorem]{Proposition}
\newtheorem{problem}[theorem]{Problem}
\newtheorem{definition}[theorem]{Definition}

\newcommand{\real}{{\mathbb{R}}}

\renewcommand{\SS}{{\mathcal{S}}}
\newcommand{\PP}{{\mathcal{P}}}
\newcommand{\RR}{{\mathcal{R}}}

\newcommand{\UU}{{\mathcal{U}}}

\newcommand{\YY}{{\mathcal{Y}}}
\newcommand{\ZZ}{{\mathcal{Z}}}

\renewcommand{\epsilon}{\varepsilon}

\newcommand{\until}[1]{\{1,\dots, #1\}}

\newcommand{\setdef}[2]{\{#1 \; | \; #2\}}

\newcommand{\TwoNorm}[1]{\|#1\|}

\newcommand{\oprocendsymbol}{\hbox{$\bullet$}}
\newcommand{\oprocend}{\relax\ifmmode\else\unskip\hfill\fi\oprocendsymbol}

\newcommand{\context}{\mathbb{C}}

\newcommand{\indexx}{j}

\renewcommand{\theenumi}{(\arabic{enumi})}

\begin{document}

\title{Classifying Emergence in Robot Swarms:  \\ An Observer-Dependent Approach}

\author{Ricardo Vega\qquad Cameron Nowzari \thanks{The authors are
    with the Department of Electrical and Computer Engineering and Department of Computer Science,
    George Mason University, Fairfax, VA 22030, USA, {\tt\small
      \{rvega7,cnowzari\}@gmu.edu}}}
       
\maketitle 



\begin{abstract}
Emergence and swarms are widely discussed topics, yet no consensus exists on their formal definitions. This lack of agreement makes it difficult not only for new researchers to grasp these concepts, but also for experts who may use the same terms to mean different things. Many attempts have been made to objectively define "swarm" or "emergence," with recent work highlighting the role of the external observer. Still, several researchers argue that once an observer’s vantage point (e.g., scope, resolution, context) is established, the terms can be made objective or measured quantitatively. In this note, we propose a framework to discuss these ideas rigorously by separating externally observable states from latent, unobservable ones. This allows us to compare and contrast existing definitions of swarms and emergence on common ground. We argue that these concepts are ultimately subjective—shaped less by the system itself than by the \textit{perception} and \textit{tacit knowledge} of the observer. Specifically, we suggest that a "swarm" is not defined by its group behavior alone, but by the process generating that behavior. Our broader goal is to support the design and deployment of robotic swarm systems, highlighting the critical distinction between multi-robot systems and true swarms.

\end{abstract}

\section{Introduction}

The lack of a clear definition of what ``emergence" or a ``swarm" is often leads to confusion and miscommunication. Thomas Kuhn \cite{TSK:97} warned about the danger of incommensurability of scientific theories, and anyone who does research in these areas and has done a deep dive search of the literature is likely already aware of these problems. Today, the word swarm has become a buzzword that is thrown around and often applied to any general multi-agent system making it difficult to not only find and understand the current state of the art but also to advance the technology as a whole. 

\begin{figure}[h]
\centering
  \subfigure[]{\includegraphics[height=.2\linewidth]{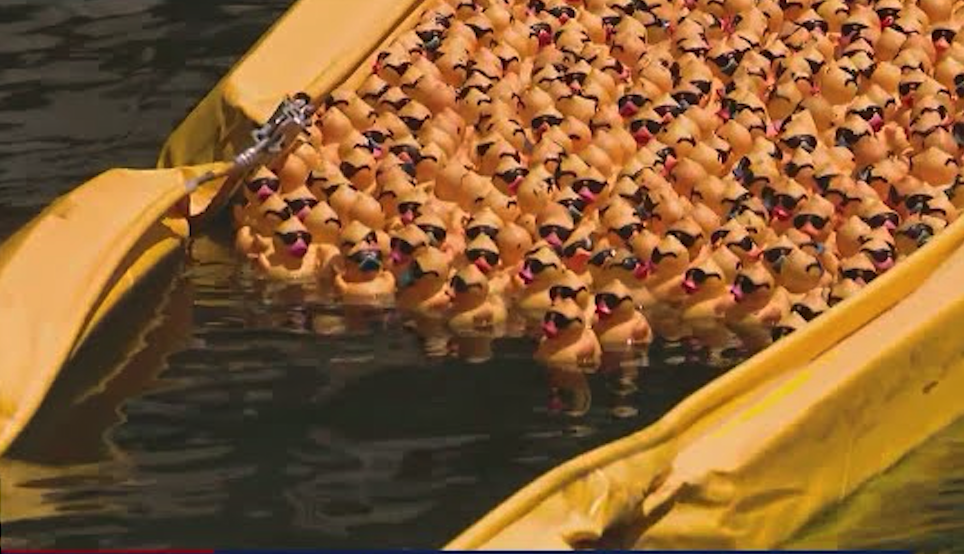}} \hfill
  \subfigure[]{\includegraphics[height=.2\linewidth]{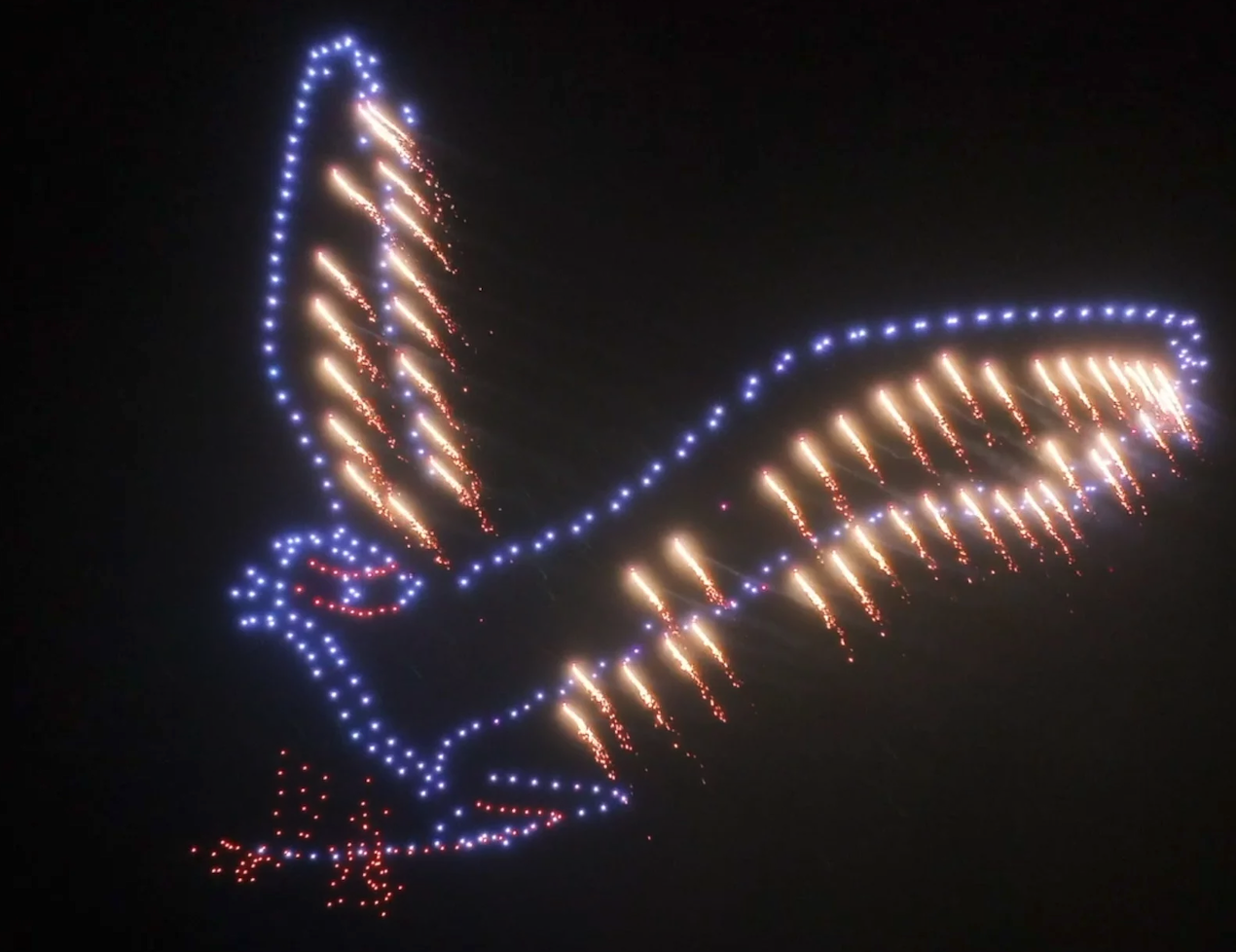}} \hfill
  \subfigure[]{\includegraphics[height=.2\linewidth]{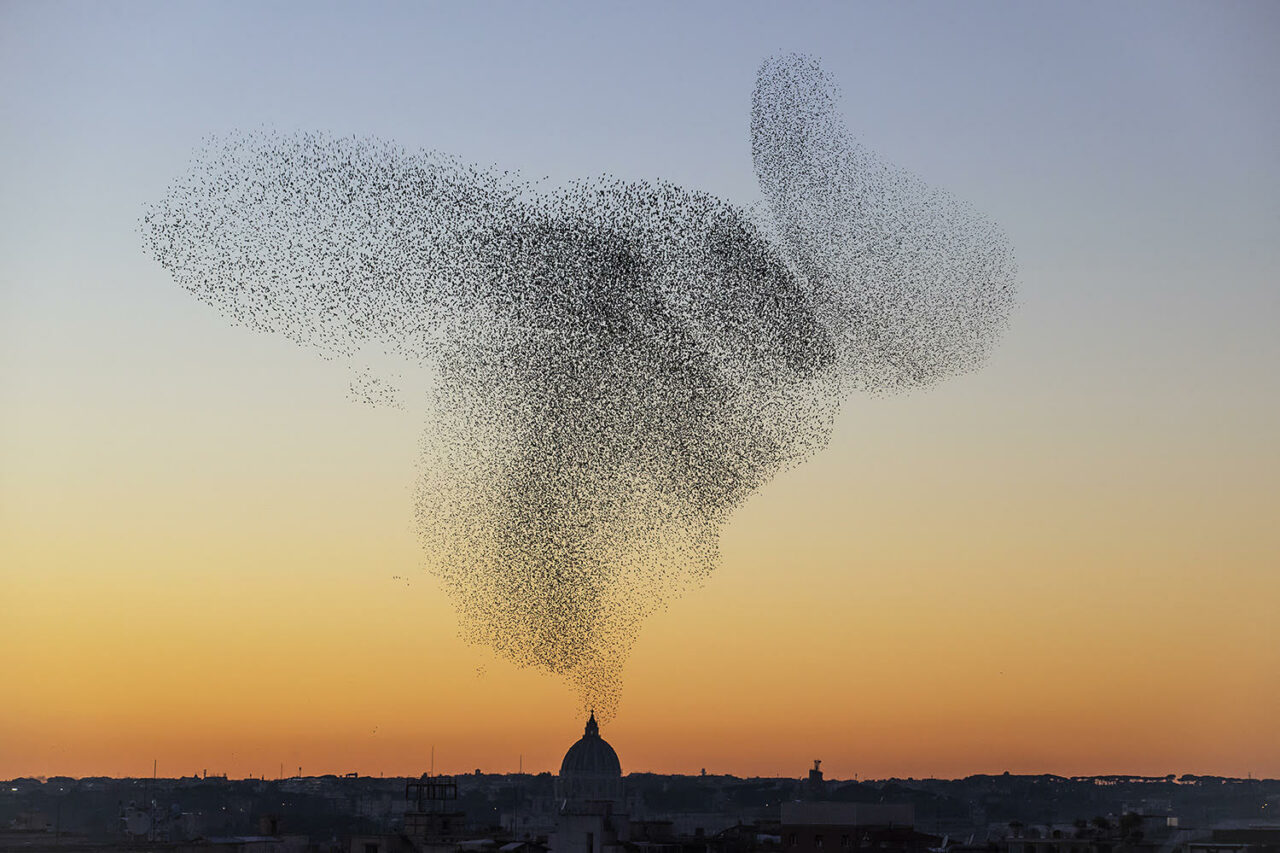}}
  \caption{Examples of candidate swarm systems including (a) The Chicago Ducky Derby, (b) a drone light show, and (c) a starling murmuration. Which of these three systems are considered a swarm?}
  \label{fig:panel}
\end{figure}

Consider the three images shown in Fig.~\ref{fig:panel} and try to determine whether each one is a swarm or not. In (a) we see a picture of the annual Ducky Derby race in Chicago, where more than 75,000 rubber ducks are thrown in the river to produce a clear flocking behavior. In (b) we see a modern day drone light show, where the image of a bird ``emerges" from the positions and LEDs on drones. In (c) we see a starling murmuration.

Most people tend to agree that (a) is not a swarm, and (c) is indeed a swarm. But when asking about (b) we find mixed responses depending on the respondent's background and expertise. Engineers tend to view any successful coordination as a swarm (e.g., the drone light show in~(b)) vs biologists/psychologists who would adamantly argue that it is not a swarm. This occurs because engineers focus generally on the emergent properties (e.g., 400 drones successfully making a bird image) whereas the scientists focus more on the \textit{process} of emergence rather than the end result. The goal of this article is not necessarily to objectively answer this question and generate consensus, but rather dig into the reasons behind the determination of a system being a swarm or not. 

Instead of arguing whether the drone show is or is not a swarm, the only way to settle this debate as a community is to acknowledge the objective distinctions between them by carefully considering the resolution, scope, and tacit knowledge of an external observer and separating resultant emergent properties from the processes that gave rise to them.

\subsection{The Incommensurability of Scientific Theories and Rediscovery of Old Ideas}
\textbf{Question:} What do the following research areas, tools, and/or methods have in common?

Emergence, Swarms, Panarchy, Morphogenesis/Morphogenetic Engineering, Cybernetics, Complex Adaptive Systems, Autopoiesis, Coleman's Boat, (Self-Organized) Criticality, Phase Transitions 

\textbf{Answer:} Quite a lot. But we are not able to find many people who can actually answer this question precisely with a clear awareness of not only their existence, but more importantly the methods and approaches within each and how they relate. A quick google search of any of these concepts should fairly quickly reveal the massive amounts of similarities, but there is very little awareness about the state-of-the-art in these different pockets of research. 
 
The issue here is one of incommensurability as not only did all these words originate at different times, for different purposes, by different groups of people, but even the definitions of the same word may be changing over time. If the umbrella term in this list is `emergence', we must acknowledge it originated in the early history of the philosophy of science, and has only been more recently adopted in more modern ideas or fields like systems theory and complexity science. With so many different groups of people with varying expertise using the same words interchangeably, it's very easy to get confused navigating the literature of swarms and controlling complex systems more generally. This has led to the different groups creating their own terms, but we believe this to now be an even bigger problem as a society, leading to rediscovery and re-branding of old ideas, most problematically due to a lack of awareness. Table~\ref{ta:emergence} presents the way we separate both the types of emergence from the process that produces them.

\renewcommand{\arraystretch}{1.6}

\begin{table}[h]
\begin{tabularx}{\textwidth}{l|X|X|X}
\toprule
\textbf{Mechanism $\backslash$ Type} & \textbf{Nominal} & \textbf{Weak} & \textbf{Strong} \\
\midrule

\textbf{Order from Disorder} & 
\textbf{Fields:} Statistical Mechanics, Thermodynamics, Chemistry \newline
\textbf{Topics:} Entropy, Temperature, Ensembles \newline
\textbf{Refs:} \cite{LB:1877, ETJ:57, HBC-HLS:98} & 

\textbf{Fields:} Condensed Matter Physics, Systems Biology \newline
\textbf{Topics:} Phase Transitions, Protein Folding, Pattern Formation \newline
\textbf{Refs:} \cite{PWA:72, SAK:92, NG-CW:11} & 

\textbf{Fields:} Quantum Foundations, Philosophy of Physics \newline
\textbf{Topics:} Decoherence, Quantim-Classical Transition, Holism \newline
\textbf{Refs:} \cite{RBL-DP-JS=BPS-PW:00, DW:12, GFRE:06}\\

\midrule

\textbf{Causality} & 
\textbf{Fields:} Classical Mechanics, Molecular Biology, Modern Control Theory \newline
\textbf{Topics:} Mechanistic Explanation, Kinematics, Deterministic Causality \newline
\textbf{Refs:} \cite{IN:1687, PD-LFA:05} &

\textbf{Fields:} Complex Systems, Ecology, Network Neuroscience, Cybernetics \newline
\textbf{Topics:} Effective Information, Feedback Loops, Agent-Based Models \newline
\textbf{Refs:} \cite{JHH:95, OS:16, SAL:98} &

\textbf{Fields:} Philosophy of Mind, Theoretical Psychology \newline
\textbf{Topics:} Downward Causation, Supervenience, Mental Causation \newline
\textbf{Refs:} \cite{DJC:97, JRS:92, JK:00} \\

\midrule

\textbf{Self-Organization} & 
\cellcolor{gray!15} \emph{(Conceptually inconsistent: self-organization implies non-trivial emergence)} &

\textbf{Fields:} Artificial Life, Computational Neuroscience \newline
\textbf{Topics:} Cellular Automata, Hebbian Learning, Synchronization \newline
\textbf{Refs:} \cite{CGL:90, MM:09, EMI:07} &

\textbf{Fields:} Consciousness Studies, Integrative Mind Theories \newline
\textbf{Topics:} Integrated Information Theory, Enactivisim, Autopoiesis \newline
\textbf{Refs:} \cite{HRM-FJV:12, GT:04, TWD:11, FJV-ET-ER:17} \\

\bottomrule
\end{tabularx}\caption{Types and Mechanisms of Emergence showing the focus points of different Fields of science. This paper focuses specifically on self-organizing robot systems and the distinction between nominal and weak emergence.} \label{ta:emergence}
\end{table}

\subsection{Relevant Works and Existing Definitions of Emergence}\label{se:literature}

\begin{figure}[h]
\centering
      \subfigure[]{\includegraphics[height=.3\linewidth]{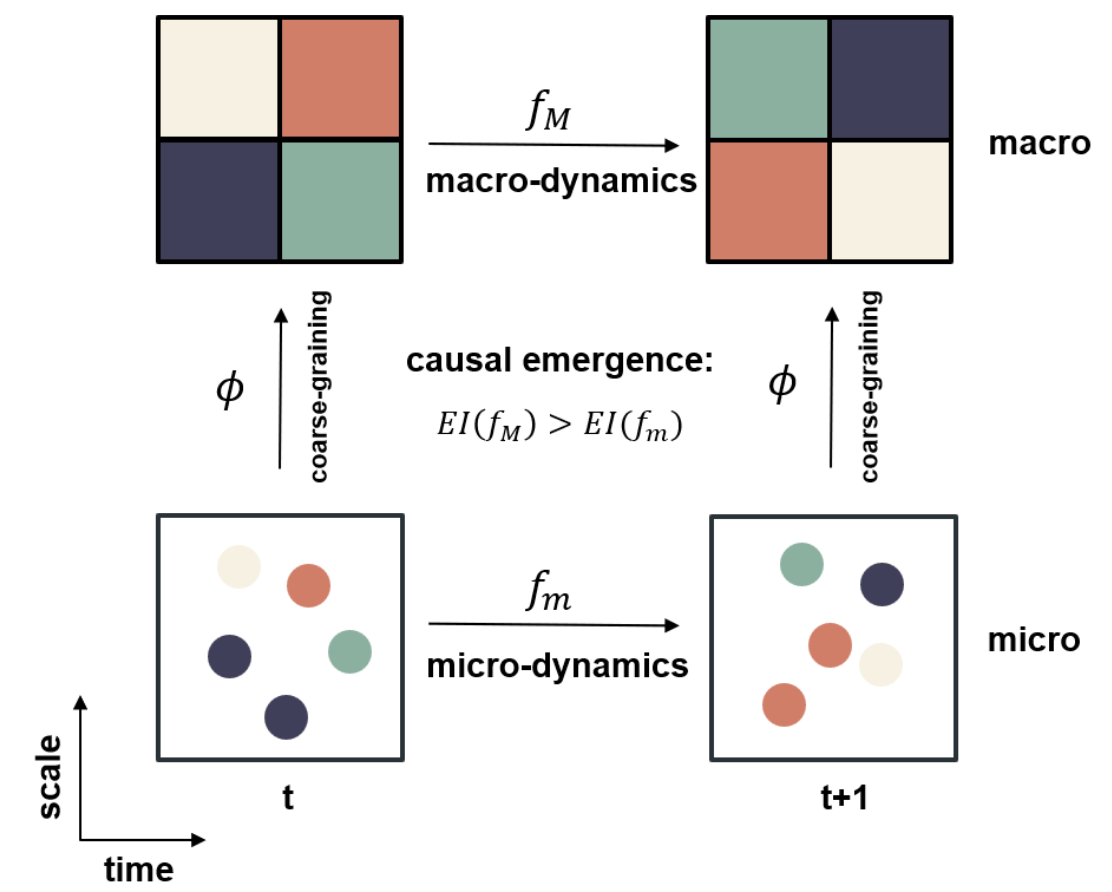}} \hfill
  \subfigure[]{\includegraphics[height=.2\linewidth]{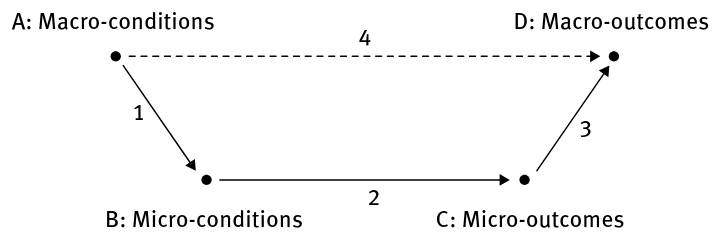}}
    \caption{Diagrams showing how the macro-level is affected by the micro-level:(a) ``The basic idea of Erik Hoel’s causal emergence theory. The colored circles represent micro-states, whereas the colored squares represent macro-states." \cite{BY-JZ-AL-JW-ZW-MY-KL-MM-PC:24} (b) Coleman's Boat shows how micro-level actions and conditions are linked to macro-level structures (and vice versa)\cite{WR-TV:17}.}\label{fig:emergence1}
\end{figure}

Emergence is a term used in many different fields such as sociology, biology, psychology, complex systems theory, and systems engineering~\cite{HK-AB-FB-BF-OH-SI:16, VD:94, SF-MF-HWC:13, RAH-NOS-GM-ES:23, BY-JZ-AL-JW-ZW-MY-KL-MM-PC:24, QL-JW-QX-YH-HZ:15, JD-YD-LM:06, RC-CC:96, BBL-BM:11}. Given the vastly varying groups of researchers and fields interested in ``emergence" who study it in different contexts, using different tools, it is very difficult to nail down exactly what ``emergence" means~\cite{PWA:72, BM:92, MAB-PH:08, MP-FB-AJR:09}. In general, emergence is defined as the creation/alteration of a property at the higher level of a system due to the interactions of the systems lower level constituents. Figures~\ref{fig:emergence1} (a) and (b) are only two representative diagrams of a very large number of similar diagrams that relate how micro (local) states and actions develop and affect the macro (group behavior) states through emergence~\cite{ MSA:95, JSC:84, WR-TV:17, GM:05,RJ-JWM:11, PA-TF-NF:08}.

Conceptually, emergence is often connected to the novelty/unpredictability of this occurrence and the surprise of the observer \cite{DJC:06, RA:04, EB:02, VD:94, EMAR-MS-MSC:99, JLC:94, SMC-AP:24} . Predicting emergence is therefore contradictory, the only way to know what will occur is to ``press play and find out" (i.e., simulate) \cite{SJ:02}. Already the idea that it is something that is ``expected" or ``unexpected" keys into the fact that there must be something/someone that can judge whether or not this is the case~\cite{SF-MF-HWC:13}. Various works emphasize the concept of a system's \textbf{observer} that decides whether or not emergence has occurred, meaning that emergence is dependent on what the observer not only sees but also knows of the system. However, some works exclude the knowledge of the observer from their definition saying that the emergent phenomena would occur regardless of the state of knowledge of the observer; conversely, if the phenomenon that occurs is not conceptually new with respect to the parts at the micro-level, it is simply resultant despite it may being a surprise to the observer~\cite{HK-AB-FB-BF-OH-SI:16}.

There have also been works that go further and split emergence into distinct types. The terms ``weak" and ``strong" emergence are frequently used although they don't always mean the same thing. Some works split emergence into either weak or strong depending on whether or not the emergent property can be explained or derived from the exact interactions of the lower-level constituents of the system. If the property can be rationalized by these lower level interactions then it is weak emergence, otherwise if it is not deducible from the lower-level interactions and instead it is fully unpredictable, it would be considered strong emergence \cite{DJC:06, MAB:97, MM-CMS:11}. Other works continue with this way of defining types of emergence by the ``surprise" of the observer and whether or not the behavior can be predicted, explained or reproduced (simple, weak, strong, spooky emergence \cite{SM-LR:15}). This suggests that things we label today as ``strongly emergent" may just be a lack of a deeper understanding of the system. This is especially problematic as one may argue that classification of a certain type of system may change over time as our understanding about it grows~\cite{CG:25}.

\paragraph*{Quantitative Measures of Emergence}
Besides conceptual models and philosophical debates, there are also many groups working to quantify emergence through mechanistic models. aiming for numerical descriptions and even closed-form equations to measure emergence~\cite{YBY:04, BY-JZ-AL-JW-ZW-MY-KL-MM-PC:24, EB-JLD:11, JPC:94, AJR:07, GSB-CG-NF:17,VD:94, AKS:08}. There have been attempts to formally define emergence using grammar and math though trying to come up with a way to model all types of emergence, and especially strong emergence, is rather difficult. Works have tried chipping away at formalizing a definition by defining concepts such as resolution, scope, variety, entropy, and redundancy~\cite{YBY:04, MM-CMS:11, AJR:07} that do allow us to think about emergence in a more mathematical way, though they usually still abstract away some things in order to make these ideas work (e.g., resolution is used to help distinguish between macro and micro state though how to actually measure resolution is not clear). 

\subsection{Scope and Intended Audience}
Table~\ref{ta:emergence} shows a very high level snapshot of what research in ``emergence" might look like. Bedau~\cite{MB:02} classified emergence into
three categories according to the causal interactions between the micro- and macro-levels: nominal emergence \cite{RH:85, NAB:94}, weak emergence \cite{MAB:97}, and strong emergence\cite{JK:92, TO:20}. We will formalize these types of emergence in Section~\ref{se:emergence} but for now we can think of them as follows:

\textbf{Nominal Emergence} - Interacting parts with a clear connection between the parts and the whole (e.g., different interacting parts of a functioning clock).

\textbf{Weak Emergence} - A clear connection between the parts and the whole but unclear mechanics / interactions between the two (e.g., starling murmurations or schools of fish). 

\textbf{Strong Emergence} - Irreducibility of the whole to the parts (e.g., consciousness emerging from a network for neurons and synapses).

While the lines in Table~\ref{ta:emergence} are beginning to blur today, the first column is where traditional top-down engineering and modern control theory sits. The rows instead separate the method in which the emergent process occurs. People try to understand emergence from different aspects, including self-organization \cite{NL:95}, order out of disorder \cite{JPC:94, JHH:00}, and causality \cite{BY-JZ-AL-JW-ZW-MY-KL-MM-PC:24}.

To help ground and distinguish this paper from the others, we focus on the specific engineering goal of wanting to design local behaviors that give rise to desired group behaviors. Specifically in the context of designing and deploying swarms of robots to solve problems, engineers desire systems that are both adaptable and flexible, but simultaneously controllable and predictable. Designing systems that don't exhibit any weakly emergent properties suggest they are inflexible, and this captures the elusiveness of designing autonomous systems that can operate in unknown or highly dynamic situations. 

The intended audience of this paper is anyone interested in designing or deploying swarms to solve real problems. More specifically, we aim to bridge the fields of modern day control systems engineering and Agent-Based Modeling (ABM) and simulation. Engineering traditionally takes a top-down approach, focusing on an end result and designing one very good method of producing that result. On the other hand, the ABM community employs a bottom-up approach, focusing on what kinds of low-level processes or interactions produce different results. 

To combine these ideas and produce practically useful systems, we believe a combination of top-down management/guidance must be incorporated with lower level ``swarms" that exhibit consistently reproducible behaviors. A single sheepdog guiding an entire herd of sheep is a perfect example of this combination actually solving a real problem~\cite{PK-SH-HV:02dynamics, PK-SH-HV:02, TN-TN:06, SR-QM-SHY:11, BB-MT:12}, but we have yet to see similarly designed/engineered systems. Towards this end, this paper deeply examines the specific mechanisms that produce various group behaviors by separating the finally emergent result of a system, from the specific manner in which it was produced. 

\paragraph*{Statement of Contributions} 
The contributions of this paper are threefold. First, we lay a formal mathematical framework to enable an objective discussion about the traditionally subjective concepts in this paper, borrowing many ideas from Swarm Analytics~\cite{AJH-AH-DJR-HAA:23}. Second, we dive into the meanings of the word ``emergence" and ``swarm" and distinguishing them as two separate, but clearly related, concepts. A few running examples will be used to help illustrate these differences. Finally, we close with some ideas for exciting and promising directions of research specifically focusing on the intersection of bottom-up self-organization and top-down engineering.

\paragraph*{Organization}

In Section~\ref{se:formal} we formalize the mathematical objects and variables needed throughout the paper. Section~\ref{se:emergence} reviews the term ``emergence" in the literature and compares/contrasts different definitions on common ground. In Section~\ref{se:swarm} we turn our attention to the term ``swarm" and how it relates to emergence. In Section~\ref{se:outlook}, we review a narrow aspect of the literature that aims precisely at bridging the ideas of this paper together and encouraging further research. Finally, we conclude with our main takeaway thoughts in Section~\ref{se:conclusions}.

\section{Formal Definitions and Methods}\label{se:formal}

A system of systems is anything that can be separated into~$N > 1$ distinct components that interact but can be described separately within a context~$\mathbb{C}$. 

We assume general and arbitrary local dynamics for each agent~$i \in \until{N}$
\begin{align}\label{eq:localdynamics}
\dot{p_i} &= f_i(p_i,u_i,w_i),
\end{align}
where~$p_i(t) \in \mathcal{P} \subset \real^{d_p}$ is the externally observable state of agent~$i$,~$u_i(t)$ is the local action, and $w_i$ captures everything external to the agent, including disturbances and interactions with the environment. 

From here we follow closely the formalism in~\cite{AJH-AH-DJR-HAA:23} with a few changes that will be discussed as needed. All the observable \textbf{data} about the system is contained in the trajectories~$P(t) = (p_1(t), \dots, p_N(t))$ and we let~$F: \mathcal{P}^N \rightarrow \mathcal{Y}$ be a map that processes the data into the set of all \textbf{information}~$\YY$ with elements~$Y = F(P)$. For simplicity and with a slight abuse of notation, we may refer to~$P$ to denote either to states at a particular instance of time but also the entire history of all trajectories.

\begin{definition}[Information Marker]
{\rm
An \textit{information marker} is a subset of information~$M = G(Y) \in \real^{m}$. In general these are reduced-order measurable outputs~$G: \mathcal{Y} \rightarrow \real^{m}$ of the entire state~$P(t)$ to summarize the information contained within the entire system (common examples include average speed or center of mass). 
}
\end{definition}

\begin{definition}[Structure]
{\rm
The \textit{structure} of a system is a subset of the information space~$\eta \subset \real^m$ that implicitly partitions the set of all trajectories~$P(t)$ into two nonintersecting regions (e.g., is the average speed of the system below a certain value or not?).
}
\end{definition}

\begin{definition}[Group Behavior]
{\rm
A \textit{group behavior}~$B_j$ is a label defined by an explicitly definable structure~$\eta^\indexx$ such that~$M^\indexx \in \eta^\indexx$ implies the trajectories of~$P(t)$ are contained to a sub-manifold of~$\PP^N$. The set of all group behaviors is $\mathcal{B} = \{B_1, B_2, \dots \}$.
}
\end{definition}

If~$M^\indexx \in \eta^\indexx$, then the system is said to exhibit group behavior~$B_\indexx$. Indeed this can be a temporary phase in general as~$P(t)$ and~$M(t)$ move in and out of~$\eta^\indexx$ as a particular behavior literally emerges and vanishes.  

\begin{remark}[Differences from~\cite{AJH-AH-DJR-HAA:23}]
{\rm
The term ``behavior" in~\cite[Definition 1]{AJH-AH-DJR-HAA:23} is not the same as what we call a group behavior~$B \in \mathcal{B}$. Instead, the term ``situation" in~\cite[Definition 4]{AJH-AH-DJR-HAA:23} seems to be used synonymously with group behavior in this paper, characterized as an invariant property of information~$\YY$ over a period of time. We have formalized the subset of data that maps to a behavior~$B_\indexx$ as~$\SS_\indexx$ in \eqref{eq:trajectories_behavior}. 
} 
\end{remark}

With a slight abuse of notation we let each~$B_\indexx(t) \in \{ 0, 1 \}$ represent whether a particular system~$P(t)$ is exhibiting behavior~$B_\indexx$ at that time or not. 
It is worth mentioning the infinite set of behaviors~$\mathcal{B}$, including both existing behaviors and ones still to be discovered, are not mutually exclusive compartments and it is possible for a system~$P(t)$ to exhibit multiple different group behaviors simultaneously~$\sum B_\indexx(t) > 1$. 

Given the original unconstrained domain of all agent states~$P \in \mathcal{P}$, the subset of states that represent a behavior~$B_\indexx$ can be implicitly defined
as the set of all trajectories~$P(t)$ that (exhibit information markers that) exhibit behavior~$B_\indexx$. Formally,
{\begin{align}\label{eq:trajectories_behavior}
\SS_\indexx \triangleq \setdef{P \in \PP}{G(F(P)) \in \eta^\indexx} ,
\end{align}} 
i.e., $M^j \in \eta^j \iff P \in \SS_\indexx$ with the dimension of~$\SS_\indexx$ smaller than the dimension of~$\PP^N$.

\begin{figure}[h]
\centering
    \includegraphics[width=.99\linewidth]{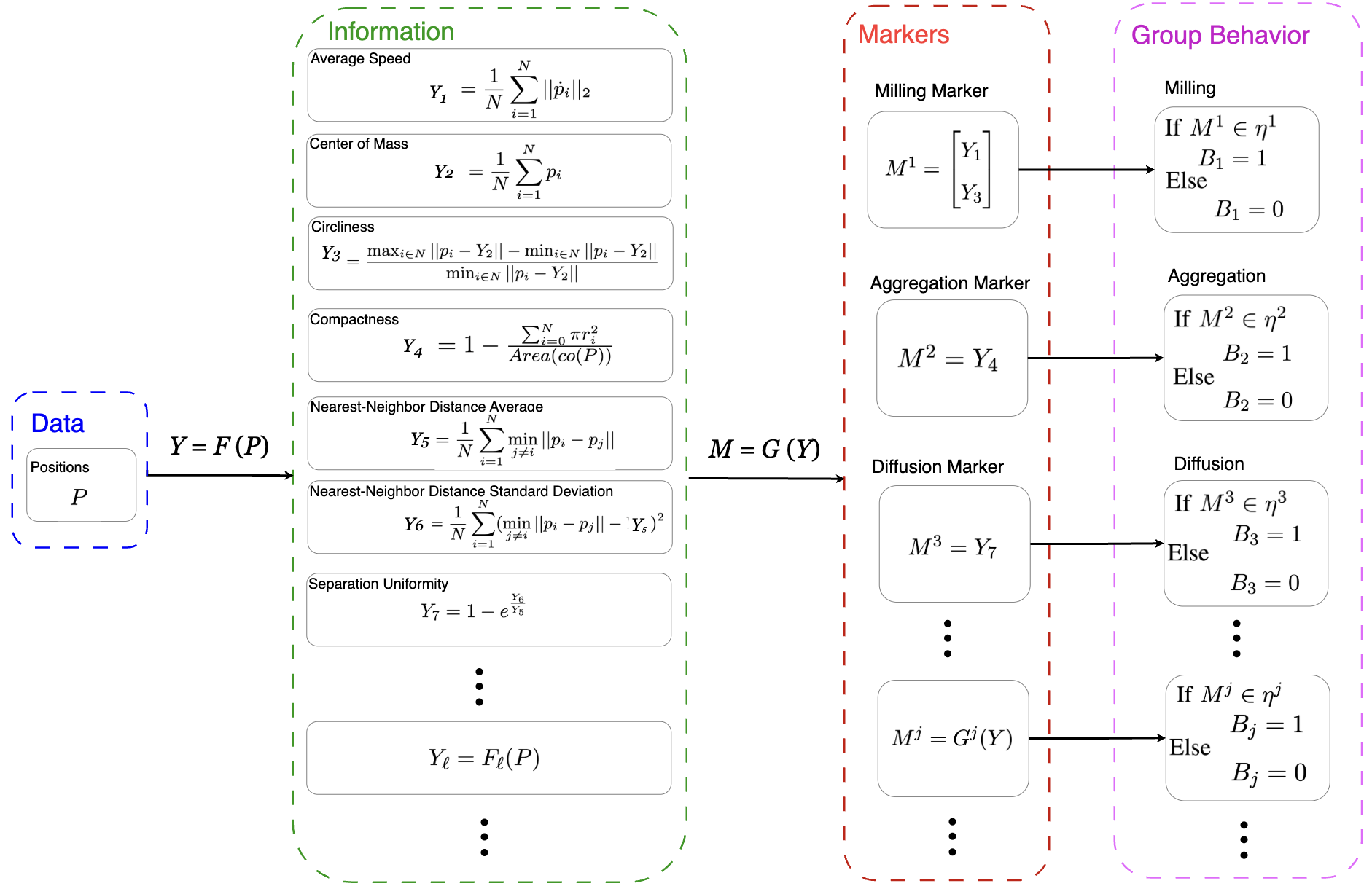}
    \caption{Observable Data to processed Information to Information Markers to Group Behavior Classification. For each behavior~$B_\indexx \in \mathcal{B}$, there exists a set of marker states~$M^\indexx$ and a subset~$\eta^\indexx$ that defines the behavior. Here, $r_i$ is the radius of agent $i$, and $\text{co}(P)$ is the convex hull of $P$~\cite{FB-JC-SM:09}. }
    \label{fig:information-behavior}
\end{figure}

Figure~\ref{fig:information-behavior} shows the steps in how observable data from a potential swarm are used to determine if a particular group behavior~$B_\indexx$ is present by mapping (1) data to information; (2) information to markers; and (3) markers to behaviors:

\begin{enumerate}
\item All observable data~$\{P(t)\}$ over some window of time is processed as information~$Y = F(P)$. 
\item For each behavior~$B_\indexx$, the relevant markers are extracted in~$M^\indexx = G^\indexx(Y)$.
\item If the marker states lie in the cluster~$M^\indexx \in \eta^\indexx$ as defined by the behavior, then we say the agents~$P(t)$ are exhibiting group behavior~$B_\indexx$ and define~$B_\indexx = 1$. 
\end{enumerate}

We now construct a very simple running example to demonstrate the main ideas of the paper. 

\begin{example}[Milling Agents]\label{ex:intro}
{\rm
You stumble across what looks like~$N = 6$ agents distributed equally around a unit circle and moving around in unison (e.g., cyclic pursuit~\cite{JM-MEB-BAF:04, ZL-MEB-BAF:04, QW-YW-HZ:16, CW-GX:17, RZ-YL-DS:15, XY-LL:17, NM-NM-AJ-KD:09, NC-MDM-AG-AG:08, YH-RA:08, RZ-ZL-MF-DS:15, EF-DAL-SM:08, SY-SP-YK:13, CT-CL-CN:20}). You don't know exactly what they are doing or what their internal dynamics are, but their trajectories over time are obvious and easy to characterize:

Let the initial 2D position of each agent be given by
\begin{align}\label{eq:observable}
p_i(0) = \begin{bmatrix} \cos \frac{i \pi}{6}  \\ \sin \frac{i \pi }{6} \end{bmatrix}, \quad i = \until{6},
\end{align}
they are then moving around the circle with unit speed and completing a full rotation every~$T = 2\pi$ seconds. Their trajectories for all time~$t \geq 0$ are then simply
\begin{align*}
p_i(t) = \begin{bmatrix} \cos (\frac{i\pi}{6} + t) \\ \sin (\frac{i\pi}{6} + t) \end{bmatrix}.
\end{align*}
Their internal dynamics~$f_i$, local controller~$u_i$, and latent states~$z_i$ are unavailable/inaccessible to an external observer. What is available are their positions~$P(t)$ and thus the resultant kinematics
\begin{align}
\dot{p_i}(t) = \frac{d p_i(t)}{dt} =  \begin{bmatrix} -\sin (\frac{i\pi}{6} + t) \\ \cos (\frac{i\pi}{6} + t) \end{bmatrix}.
\end{align}

With access to all available data, the first step of Fig.~\ref{fig:information-behavior} is to turn this data into measurable information. Due to the simplicity of our exactly circular trajectories~$P(t)$ shown in Figure~\ref{fig:milling}, we can easily read out some of the information states like average speed~$Y_1 = 1$, center of mass~$Y_2 = (0, 0)$, or circliness~$Y_3 =0$.

Let us now consider what ``group behaviors" this ``swarm" of~$N=6$ agents looks to be exhibiting. Let us consider the 3 group behaviors shown and defined in Fig.~\ref{fig:information-behavior}: Milling ($B_1$), Aggregation ($B_2$), and Diffusion ($B_3$). To evaluate the existence of behavior~$B_j$, we must consider only the information~$M^j \subset Y$ and see if it belongs to the set~$\eta^j$. 

The ``milling" behavior~$B_1$ is defined by its average speed~$Y_1$ and circliness~$Y_3$. The information markers relevant to milling are thus
\begin{align*}
M^1 =  \begin{bmatrix} Y_1 \\ Y_3 \end{bmatrix} = \begin{bmatrix}  \frac{1}{N}\sum_{i=1}^N \TwoNorm{\dot{p}_i}_2  \\ \frac{\max_{i \in N} ||\mathbf{p}_i-Y_2|| - \min_{i \in N}||\mathbf{p}_i-Y_2||}{\min_{i \in N}||\mathbf{p}_i-Y_2||} \end{bmatrix} = \begin{bmatrix} 1 \\ 0 \end{bmatrix}. 
\end{align*}

The structure set~$\eta^1$ requires that the average speed be nonzero~$Y_1 > 0$ and the circliness~$Y_3 = 0$. Since the trajectories~$\{P(t)\}$
have been purposely constructed to achieve a perfect ``circliness" score~$Y_3 = 0$ and the average speed of agents is nonzero, we conclude that this system is indeed exhibiting the milling behavior and thus~$B_1 = 1$. 
a
The ``aggregation" behavior~$B_2$ is defined by its compactness~$Y_4$,
\begin{align*}
    M^2 = Y_4 = 1 - \frac{\sum_{i=0}^N \pi r_i^2}{Area(co(P))} ,
\end{align*}
where $r_i$ is the body radius of agent $i$ and co$(P)$ is the convex hull of $P$ \cite{FB-JC-SM:09}. 
Essentially, compactness is a measure of how tightly the agents are huddled together. The structure set~$\eta^2$ requires this compactness to be $Y_4 = 0$ (we assume here the agents cannot overlap which would make the area of the convex hull less than the cumulative body area). In this example the area of the convex hull is much greater than the cumulative body area so therefore the system is not aggregating and thus $B_2 = 0$

The ``diffusing" behavior~$B_3$ is defined by its separation uniformity $Y_7$:
\begin{align*}
    M^3 = Y_7 = 1 - e^{\frac{Y_6}{Y_5}},
\end{align*}
where $Y_5$ is the average distance between each agent and its nearest neighbor and $Y_6$ is the standard deviation. The information can be seen as how evenly spaced the agents are to each other such that if they are perfectly spaced out, they are exhibiting the ``diffusing" behavior. The structure set~$\eta^3$ requires $Y_7$ = 0.  In this example of agents distributed equally around a circle, because they are evenly spaced out we can determine that it is exhibiting the diffusing behavior and thus $B_3 = 1$. Note in this case that the behaviors~$B_1$ (milling) and~$B_3$ (diffusal) are not mutually exclusive and in fact happen to appear together.

 \begin{figure}[h]
 \centering
     \includegraphics[width=.3\linewidth]{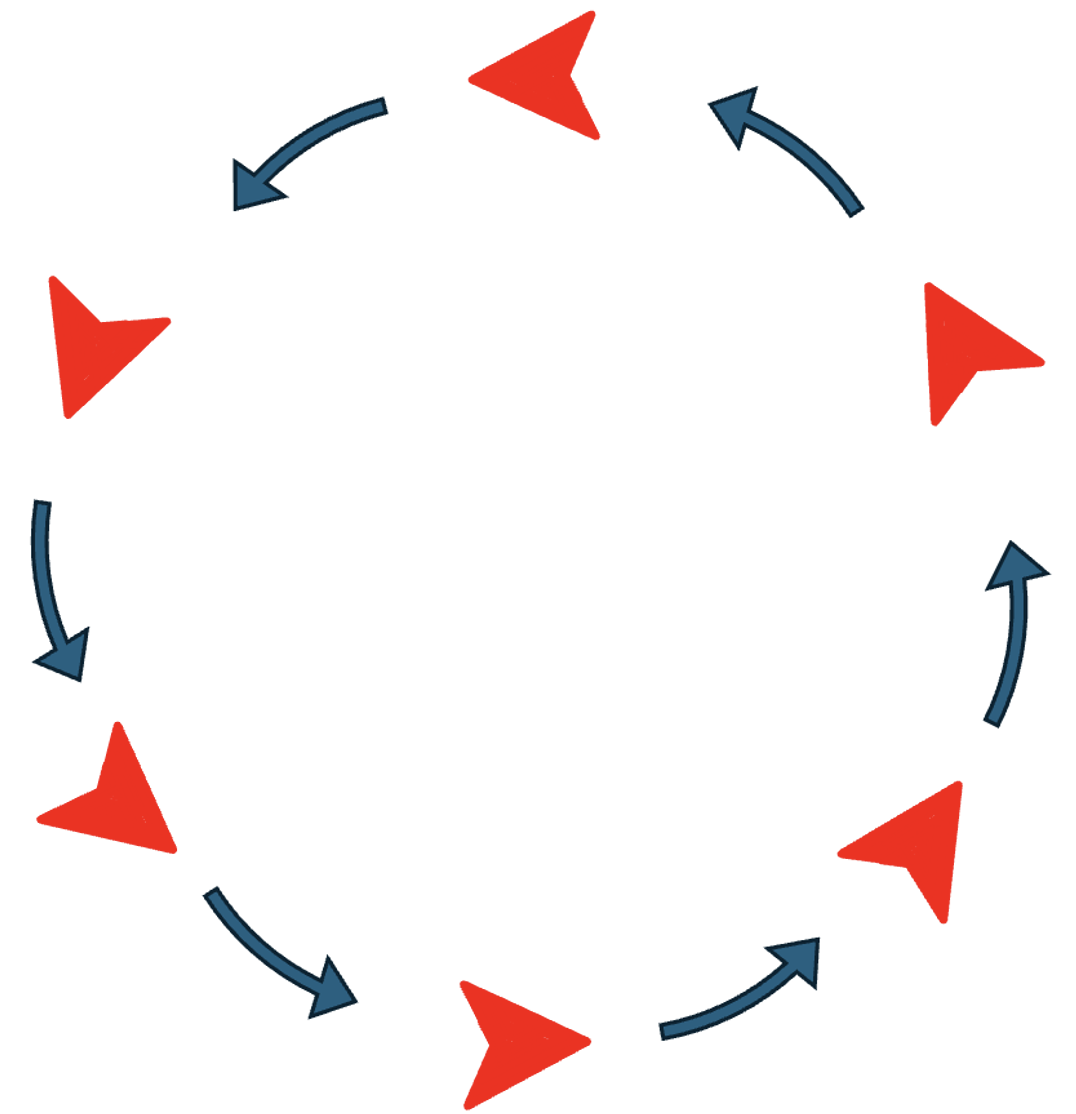}
     \caption{6 agents moving in a counter-clockwise circle.}\label{fig:milling}
 \end{figure}

}
\end{example}

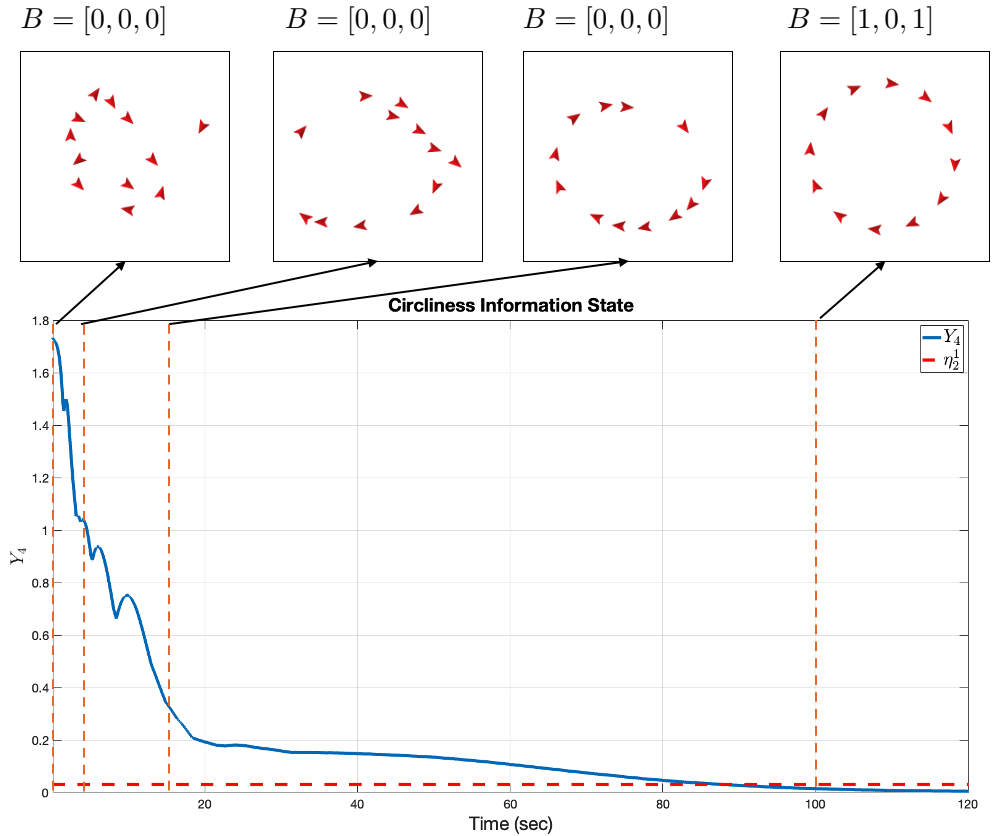
\begin{figure}[h]
\centering
  \includegraphics[width=.80\linewidth]{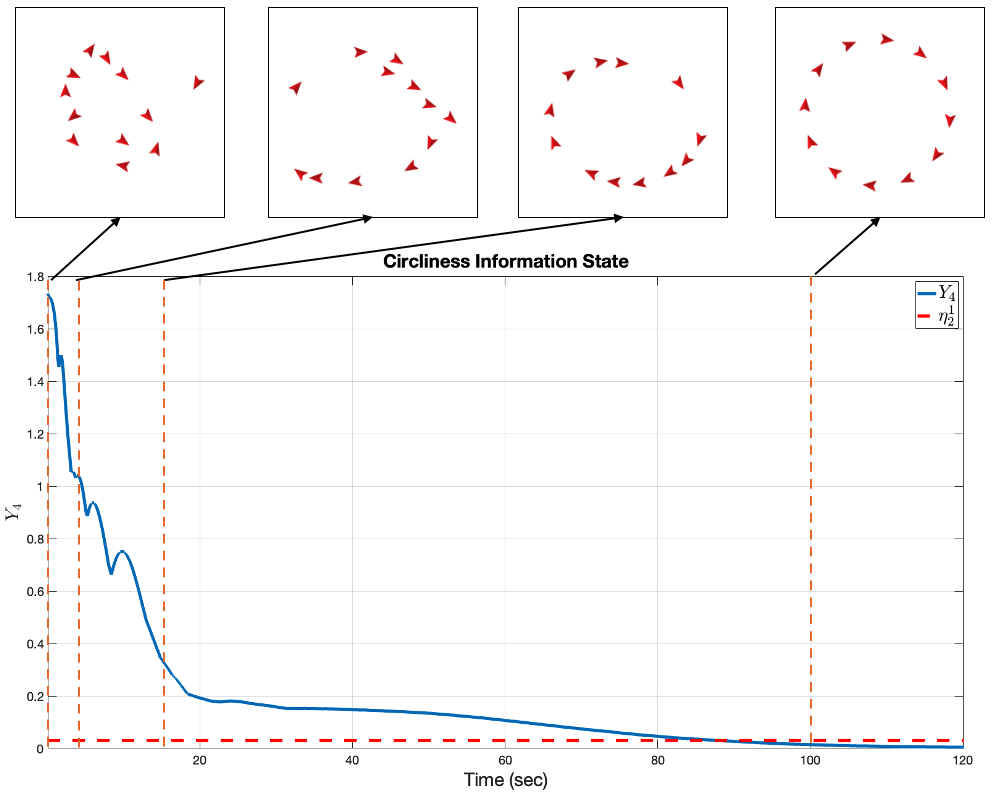}
  \put(-370,307){$B = [0, 0, 0]$}
  \put(-270,307){$B = [0, 0, 0]$}
  \put(-180,307){$B = [0, 0, 0]$}
  \put(-80,307){$B = [1, 0, 1]$}
  \caption{Plot of circliness $Y_3$ crossing the $\eta_2^1$ threshold into the milling behavior state with snapshots of system throughout run.}
  \label{fig:simulated_markers}
\end{figure}

\begin{remark}[Dynamics of Group Behaviors]
{\rm
It is worth noting that in general the information markers will be time-varying and consequently the existence of group behaviors can indeed change over time. In Fig.~\ref{fig:simulated_markers} we show a much complicated case in which the initial conditions of the agents are initially random but eventually converge to the perfect milling behavior. In this case we see that the Milling behavior does not actually emerge until about 100 seconds into the simulation and out framework is general enough to capture this. Note in this case the Diffusion behavior also emerges $B_3(t) = 0$ while Aggregation still does not and thus~$B_2(t)  = 0$ for the duration of the simulation.
}   
\end{remark}

\section{Emergence as a Process}\label{se:emergence}
With the formal definitions now in place we would like to be able to objectively answer the two questions:

\begin{problem}[Does the system exhibit emergence?]\label{pr:emergence}
Given the full trajectories of a system of systems~$P(t)$ over some time window~$t \in [0, T]$, what type of emergence does it exhibit?
\end{problem}

\begin{problem}[Is the system a swarm?]\label{pr:swarm}
Given the full trajectories of a system of systems~$P(t)$ over some time window~$t \in [0, T]$, is the system a swarm? 
\end{problem}

The remainder of this paper aims to answer these two questions objectively, both generally and specifically for Example~\ref{ex:intro}. But to first complicate matters further, we first argue why not enough information is provided to answer either. 

\subsection{Observable vs Unobservable Information and Context}

Until now, all the variables defined have been externally observable. Note that while the trajectories~$P(t)$ (and thus the kinematics~$\dot{P}(t)$) are contained in the information~$\YY$, the underlying dynamics and functional form of~$f_i$ in \eqref{eq:localdynamics} are usually not. The agents may have some internal state $z_i$ that can't be known to an observer, such as memory or computing processes, that still affect the system $X$. It is worth highlighting here that we have not explicitly made any mention of a ``sensor" or what a particular agent can perceive about the environment or world. While guesses could be made based on physical appearances of things or prior experience with similar systems, technically the sensing model and especially their internal perception models are inaccessible to an external observer. 

Besides the internal/hidden states within the agents, there are other aspects of the system, not in the observable or unobservable states, that give additional details about the system~$\mathcal{R}$. In the context of a simulation, this would be everything not captured by the state variables, including all parameters and functions describing the environment and/or interaction rules.

\begin{definition}[Context~\cite{AJH-AH-DJR-HAA:23}]
{\rm
    A context~$\mathbb{C}$ is the effective superset of information~$\YY$ required by a system and its environment to operate autonomously, and is separated into its observable, unobservable, and parameter subspaces~$\mathbb{C} = \PP^N \times \ZZ \times \RR$, respectively. 
    }
\end{definition}

\begin{remark}[How emergent is it?]
{\rm
A group behavior~$B_\indexx$ is thus anything that can be explicitly defined through a set of~$m^\indexx$ information markers~$M^\indexx \in \real^{m^\indexx}$ and a subset~$\eta^\indexx$. Trivially, the set~$\eta^\indexx$ might be very large, in which case knowing that a system exhibits a particular behavior~$B_\indexx$ might not mean much. On the other hand, for tighter definitions and smaller sets~$\eta^\indexx$, there is more constraint/order and less entropy among the agents exhibiting that group behavior. This begins to enter the realm of quantitative measures of emergence and causality that examines how predictive the future of~$P$ or $X$ might be based on their histories and for further reading on this we refer to~\cite{YBY:04, BY-JZ-AL-JW-ZW-MY-KL-MM-PC:24, EB-JLD:11, JPC:94, AJR:07, GSB-CG-NF:17,VD:94, AKS:08}. 
}
\end{remark}

\subsection{Formalized Definitions and Classifications of Emergence}

With full knowledge of how agents in a system interact, we can further define and distinguish the various types of emergence. Remarkably, we will show that all these definitions are mostly dependent on their internal (often hidden) dynamics~\eqref{eq:localdynamics}. 

\paragraph*{Type 0 - Non-emergent}
For completeness, we begin by defining systems that don't exhibit emergence trivially as any context~$\context$ in which there is only~$N=1$ thing similar to~\cite{OTH:07, YBY:04}. Formally, given the closed system context~$\mathbb{C}$, there is only one object~$N=1$.

\paragraph*{Type I - Nominal Emergence}
Type I emergence contains no feedback at all, only feedforward relationships. There may be many things~$N>>1$ but with clear function and/or rigid coordination/formation that occurs as a result of external instructions and planning. Formally, the resultant local state change does not depend on interactions with others~$\dot{p}_i = f_i(p_i, u_i, t)$ (open-loop / feedforward only).

\paragraph*{Type II - Weak Emergence}
Type II emergence is ``simple" feedback: either positive or negative. In either case, the emergent properties may be stable or unstable. Formally, the resultant local state change depends on interactions with others~$\dot{p}_i = f_i(P,u_i,t)$. It is at this level where we may start observing the phenomenon of  \textit{self-organization}, or the process by which individuals create global behaviors through just the interactions amongst themselves rather than through external intervention or instruction \cite{DW:06}.

\paragraph*{Type III - Strong Emergence}
Type IV emergence is unpredictable by definition, in that the emergent behavior cannot be reduced to the parts that gave rise to it. Formally, the context is changing in time~$\mathbb{C}(t)$ with objects being both created and destroyed, including different swarm behaviors~$B_\indexx$.

\paragraph*{Other Types of Emergence}
As mentioned earlier, there is no consensus on these definitions and other works attempt to further distinguish aspects of emergence or even partition these same types presented above differently. For example, Fromm proposes an additional type of emergence between ``Weak" and ``Strong" called `Multiple Emergence"~\cite{JT:05,JT:05ten}. 

Multiple emergence is characterized by increasing complexity in the form of multiple feedback loops, and even learning and adaptation of agents. Our view of this type of emergence is the acknowledgment that in many (especially real-world) systems, there is not only one type of emergence at play. Indeed, many different interacting systems that would otherwise exhibit their own type of emergence in isolation, can also introduce their own emergent effects in a fractal-like manner.

\subsection{Importance of Observer}
It is fully on the observer to not only identify when emergence has occurred but also to determine the type of emergence. This doesn't just include the observable information of the given system but also any beliefs/knowledge the observer has a priori. This means that given the same exact system at the same exact time, two observers can disagree on the type of emergence or even whether or not emergence occurred. 

In order for an observer to make this determination, they must be able to either observe the system at various instances in time or scale or have insight on the system itself. If only a snapshot of the system is available, the observer may only infer on the type of emergence by applying the knowledge or context they already have. If they have no relevant knowledge, they would need to see the system advance over time (e.g., a video) or observe the system at various perspectives of scale (e.g., zooming out), or both. By seeing different snapshots of the system, the observer can compare them and decide if an emergent property is present.

\newcommand{\resolution}{{\Gamma}}
\newcommand{\scope}{{\Psi}}

\paragraph*{Resolution and Scope of an Observer}
It is generally established that emergence is tied to the resolution $\resolution$ (``the finest distinction that can be
observed between two alternative system configurations") and scope $\scope$ (``set of components within the boundary between the associated system and its environment") at an observer's perspective~\cite{AJR:07}. 

For any~$N(\scope,\resolution) = 1$ (i.e., only one observable component at the perspective with $\resolution$, $\scope$), there exists~$\resolution' < \resolution$ such that~$N(\scope,\resolution') > 1$. That is, without changing the scope of observation, increasing resolution allows us to discern that a table, a brick, or a water molecule is composed of interacting or rigidly connected subatomic components. For any~$N(\scope,\resolution) = 1$, there also exists~$\scope' > \scope$ such that~$N(\scope',\resolution) > 1$. That is, anything can be seen as a smaller part of a larger system by zooming out, meaning that even determining how many agents~$N$ exist is entirely dependent on both resolution~$\resolution$ and scope~$\scope$ of an observer. Resolution and scope both have spatial and temporal components but we do not enter into this here.

While \cite{AJR:07} claims that emergence depends on resolution and scope \textit{only}, we argue that the observer's knowledge of the true inner workings of the system is also critical to determining the type of emergence or whether emergence occurs at all. 

\begin{figure*}[h]
\centering
    \includegraphics[width=.9\linewidth]{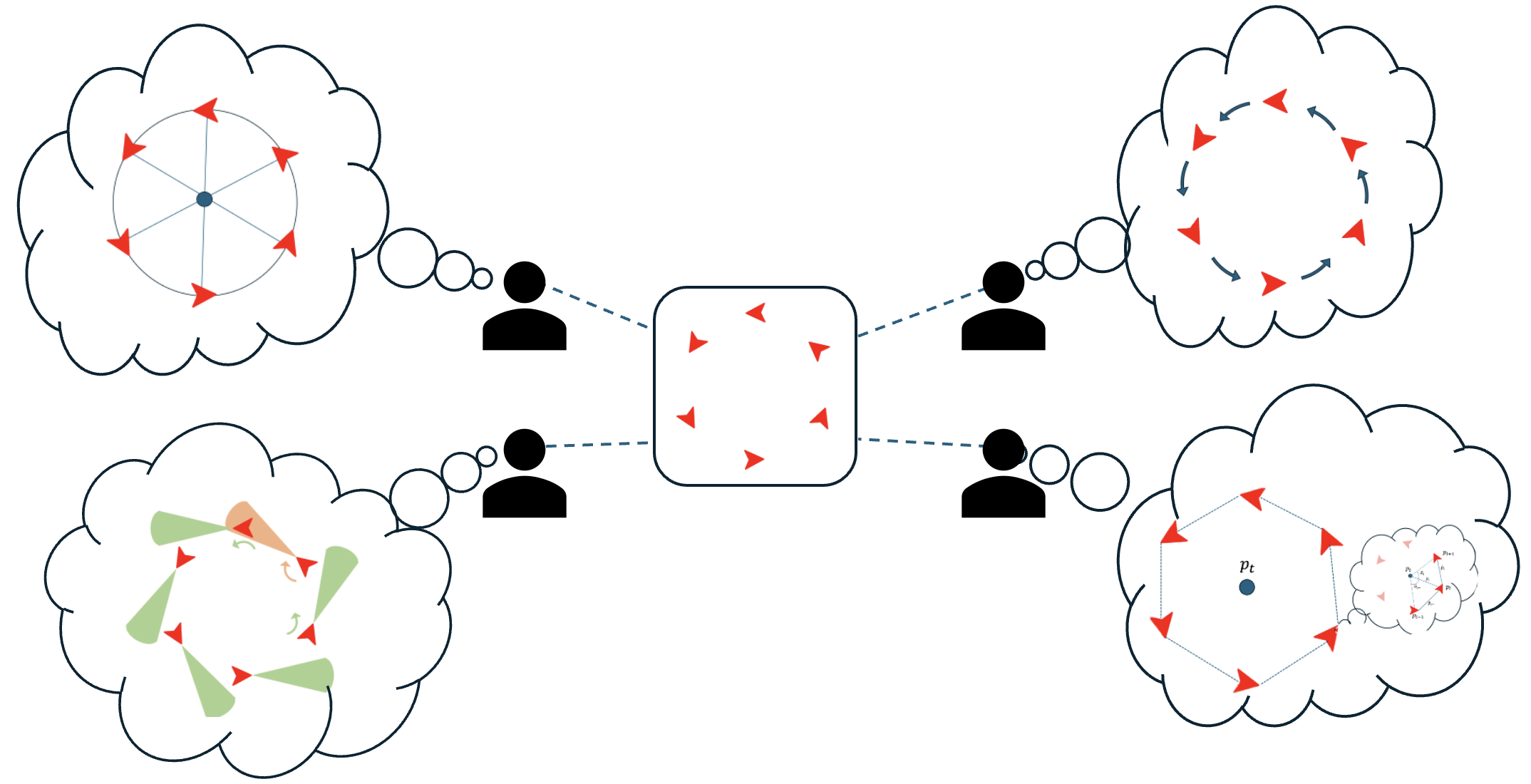}
    \put(-285,150){$(a)$}
    \put(-160,150){$(b)$}
    \put(-285, 100){$(c)$}
  \put(-160,100){$(d)$}
    \caption[Importance of the Observer's Knowledge]{Type of Emergence is dependent on perception of internal dynamics. The observer in (a) believes that the agents are all part of a single rotating part rather. The observer in (b) believes the agents are not reacting or interacting with one another and simply subject to environmental forces or on predetermined paths. In (c), the observer believes the circle is formed through the self-organization of agents that react to one another. Finally, in (d) the observer believes the agents are more aware and are intentionally making the shape by distancing themselves from a center and from their neighbors. }\label{fig:perceptions}
\end{figure*}
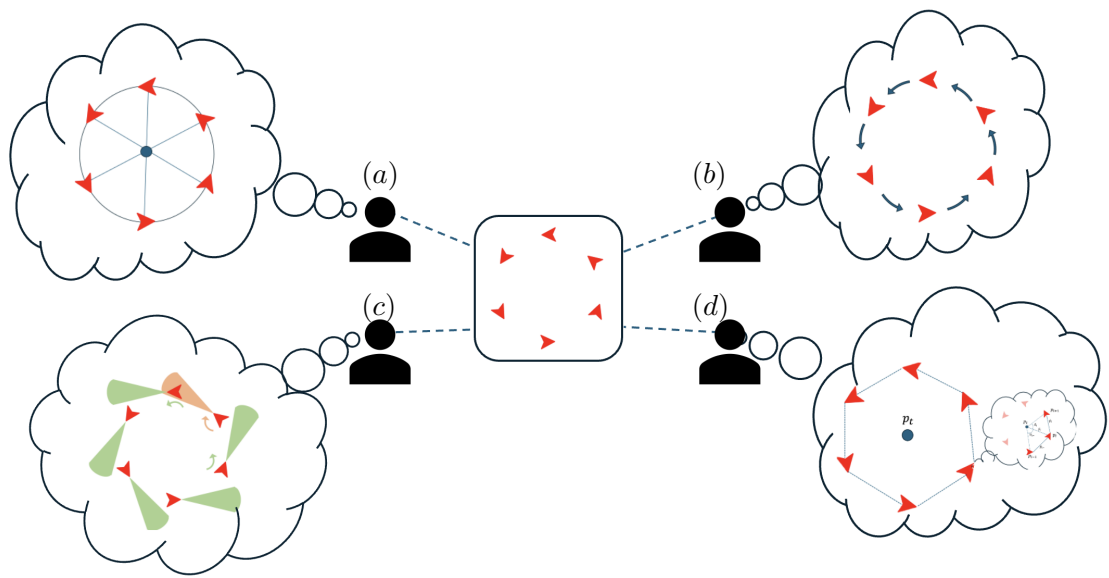

\begin{example}[Is the milling behavior emergent?]\label{ex:emergence}
{\rm Let us now revisit Example~\ref{ex:intro} and see what the answer to Problem~\ref{pr:emergence} is and how it might change depending on the particular context~$\mathbb{C}$.

\subsubsection{Apparent Milling (Type 0 - No Emergence)}

 As supported in~\cite{AJR:07}, the lack of emergence might be revealed by increasing specifically the resolution~$\resolution' < \resolution$ to realize that what originally looked like~$N = 6$ separate things, was in fact all connected to make a single rigid object~$N' = 1$.  Fig.~\ref{fig:noemergence}(a) shows a reality of what looked like~$N=6$ agents in a different resolution/scope similar to seeing only the ducks in (b) without the greater context of the merry-go-round. Since there is only a single rigid object in rotation, this is not characterized as an emergent process (i.e., Type $0$). Furthermore, even without changing the resolution (or scope), an observer may identify that the system of circling agents is just a single object based on their preconceived notion of what they are seeing.

\begin{figure}[h]
\centering
    \subfigure[]{\includegraphics[width=.3\linewidth]{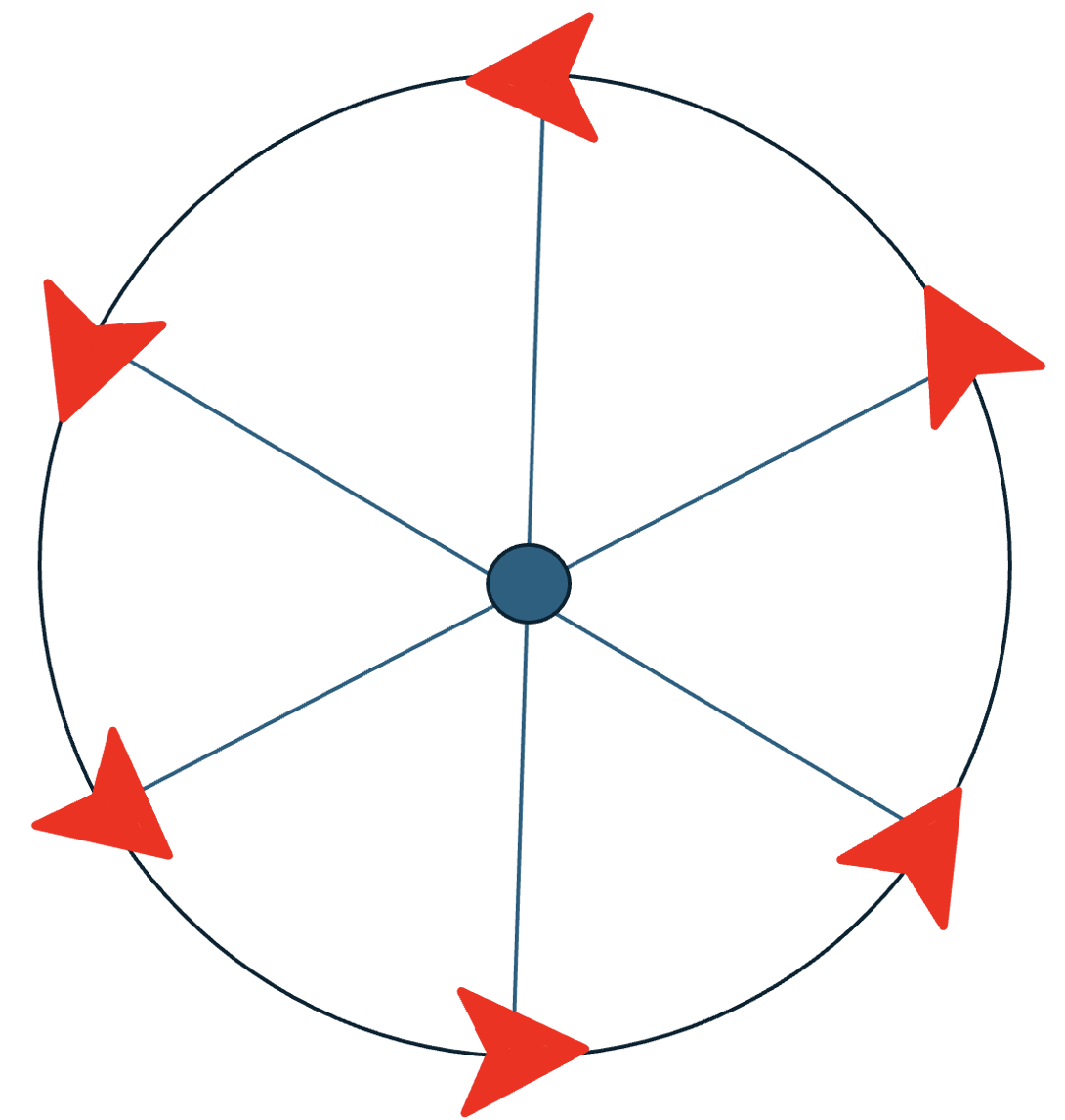}} \hspace*{6ex}
    \subfigure[]{\includegraphics[width=.3\linewidth]{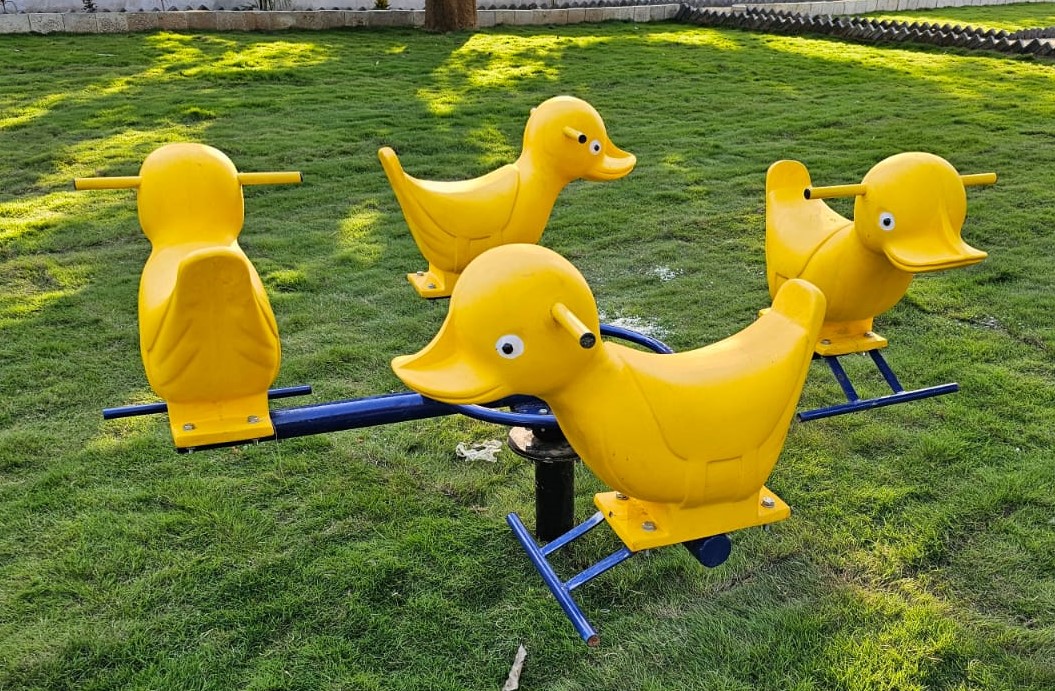}}
    \caption{A change in scope/resolution reveals in~(a) what looked like 6 independent agents as parts of a single object, similar to seeing only the ducks in (b) without the greater context of the merry-go-round.}\label{fig:noemergence}
\end{figure}

\subsubsection{Milling with no feedback (inanimate/feedforward agents) (Type I - Nominal Emergence)}
Let us now leave the resolution and scope fixed including~$N=6$ independent agents, with the additional knowledge that they have no agency~$d_u = 0, u_i = \emptyset$, the actual dynamics may be given by
\begin{align*}
\dot{p}_i = f(p_i, w_i) = \begin{bmatrix} -p_{i,2} \\ p_{i,1} \end{bmatrix} ,
\end{align*}
meaning the agents are not at all interacting and simply subject to similar environmental forces. 

Since~$\dot{p}_i$ does not depend on any other agent states~$p_{-i}$, this is generally characterized as simple or nominal emergence (i.e., Type I). Similar to the rubber ducky race in Fig.~\ref{fig:panel}(a), although there is some nominal form of `emergence', the lack of any form of feedback or agency indicates there is no self-organization.

\subsubsection{Milling with simple feedback (reactive agents) (Type II - Weak Emergence)}

Consider now Dubins' vehicle dynamics where there is a 3rd state~$p_{i,3} = \theta_i$ capturing the 2D orientation of each agent and the actual dynamics are
\begin{align}\label{eq:dubins}
\begin{bmatrix} \dot{p}_{i,1} \\ \dot{p}_{i,2} \\ \dot{p}_{i,3} \end{bmatrix} = f_i(p_i,u_i,w_i) = \begin{bmatrix} \cos p_{i,3} \\ \sin p_{i,3} \\ u_i \end{bmatrix} .
\end{align}

There now exists a single control input~$d_u = 1$ with~$u_i \in \UU = [-\omega_\text{max},\omega_\text{max}]$. This means we must also introduce a local sensor for each agent~$i \in \until{N}$. Remarkably we have avoided the concept of a sensor throughout this entire paper but it is a necessary condition to enable local feedback. 

To keep this as simple as possible, we consider only a binary sensor output~$h_i(t) \in \{0,1\}$ for each agent at any given time,
{
\begin{align}\label{eq:output}
    h_i(t) = \begin{cases}
                 1 & \text{if } \exists j \neq i, s.t. ~p_j \in \operatorname{FOV}_i , \\
                 0 & \text{otherwise,}
             \end{cases}
\end{align}
}
where $FOV_i(p_i) \subset \PP$ is the field of view of agent $i$ composed of the conical area in front of the agent with vision-distance $\gamma$ and an opening angle $\phi$~\cite{MG-JC-TJD-RG:14, DB-RT-OH-SL:18, DS-CP-GB:18, FB-MG-RN:21} as shown in Fig.~\ref{fig:milling_figure_detail}.

\begin{figure}[t]
    \centering
    \includegraphics[width=0.7\linewidth]{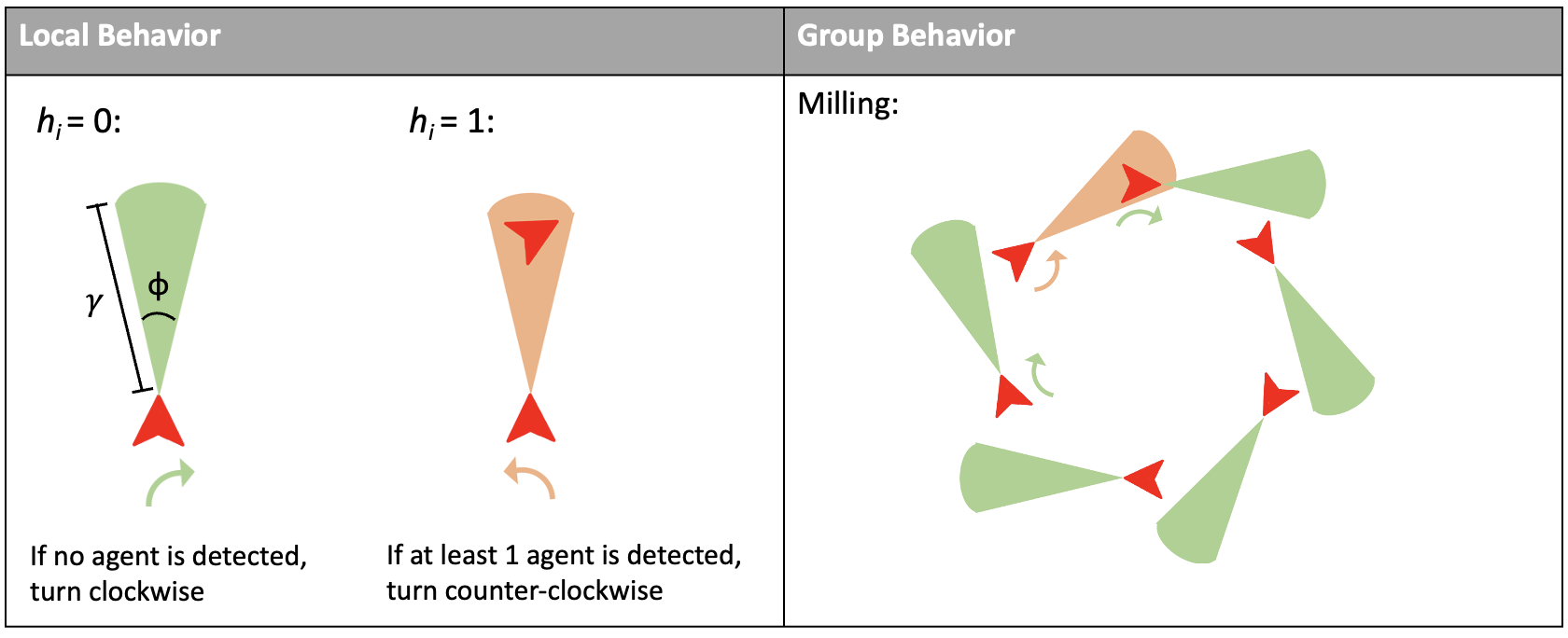}
    \caption{Given the proper conditions, reactive agents using simple feedback controller can lead to a milling swarm behavior \cite{RV-KZ-CM-DSB-CN:24, RV-CM-DSB-CN:23, MG-JC-TJD-RG:14, FB-MG-RN:21}. }
    \label{fig:milling_figure_detail}
\end{figure}

\begin{equation} \label{eq:basic_controller}
\begin{split}
u_{i}(t) &= \begin{cases} \omega_\text{max} \quad &\text{if } h_i(t) = 1 ,\\ -\omega_\text{max} \quad &\text{otherwise.} \end{cases}
\end{split}
\end{equation}

Since agents now have local agency and there is a static (memory-less) feedback law~$k : H \rightarrow \UU$, this milling circle can be characterized as weak emergence (i.e., Type II). This is the level of emergence at which the term ``swarm" usually enters the picture. It generally requires a perception of self-organization that does not exist at the lower levels. Most importantly, there is now local feedback with the introduction of local sensing and control.

\subsubsection{Milling with distributed control (cognitive agents) (Type II - Weak Emergence)}

There are a myriad of works dedicated to multi-agent cyclic pursuit or more generalized shape/formation control 
\cite{HL-ZYH-AH:21}. We borrow the term cognitive agents~\cite{AJH-ASMH-DJR-HAA:24} to refer to these but as we show here it may not be that simple. 

Consider the method proposed in \cite{CW-GX:17} that drives anonymous agents to create a circular formation with a predetermined target as the center. We don't go into the specific technical details of how this controller works, but unlike the above methods the difference here is that the low level controller of the agents are designed specifically to produce the milling behavior. Summarily, by using the relative position between agent $i$ and the target as well as between it and its neighbors, a CCW-rotating circle can be formed using the controller

\begin{align}
    u_i(t) = u^p_i(t) \beta_i(t),
\end{align}

where $u^p_i(t) \in \real^2$ is a limit-cycle oscillator that pushes the agents to rotate around the target positioned at $p_t \in \real^2$ and $\beta_i(t)$ is the component designed to ensure the angular distance between the agents is maintained.
}
\end{example}

Through these examples we can see clearly how the exact same system as viewed by an external observer might be classified differently according to the specific method in which the milling behavior was produced. We are now ready to turn to Problem~\ref{pr:swarm}.

\section{What Constitutes a Robotic Swarm?}\label{se:swarm}
With the different types of emergence now defined, we are finally ready to address the second main question this paper investigates: Given a system, within a context~$\mathbb{C}$, and its trajectories over some time~$P(t)$, can we determine if it is a swarm? Our answer is no, additional information about the unobservable states of the system would also be required.

 As we did with emergence, we again start by looking at a small cross section of common definitions of ``swarms" collected in Table~\ref{tbl:Swarm_defs}.
 The subjectivity and lack of consensus on a definition should already be quite clear. To be fair to the researchers in the table, the focus of most work is not to define ``swarm" but rather develop new methods of producing swarming behaviors. But this is exactly what this paper seeks to address, which is to enable researchers to properly explain what they are doing without conflating two things that look similar (or even identical) but have entirely different inner workings.

\renewcommand{\arraystretch}{1.3}

\begin{table}[htbp]
\centering
\caption{Commonly shared definitions of a swarm.}
\footnotesize
\begin{tabular}{|>{\raggedright\arraybackslash}p{9.1cm}|>{\raggedright\arraybackslash}p{2cm}|}
\hline
\rowcolor{gray!30}
\textbf{Definition} & \textbf{Sources} \\
\hline

1: A swarm is a group of three or more decentralized entities that display coordinated or cooperative behavior, often with limited or no centralized control. 
& \cite{MR-CA-RN:12, HH:18, AH-AS:22, RA-KC-BA-CK:19, GB-JW:93} \\

\hline
2: A swarm exhibits collective behavior that emerges from local interactions among individual agents and their environment, rather than from centralized control.
& \cite{JT:05, MS-DGR-MB:24, MB-AL-DB-MB-GF-LG-DGR-KH-MK-JK:19, ES:04, HA-SM:25} \\

\hline
3: A robot swarm is a highly redundant, decentralized, and self-organized system composed of many autonomous entities 
& \cite{AL-MB:18, AK-PW-NC-KS-ML:15, ES:04} \\

\hline
\end{tabular}

\label{tbl:Swarm_defs}
\end{table}

Everyone has a preconceived notion of what a swarm is, although defining it in a way that is clear enough to guarantee it cannot be misconstrued is clearly challenging. Each of the definitions in Table~\ref{tbl:Swarm_defs} can be used to interpret what a swarm is intuitively, however, they are not rigorous/thorough enough to objectively define it. For example, a general definition that is commonly used is that a swarm is ``a multi-agent system that cooperates to achieve a group task". However, this definition would consider a team of basketball players trying to win a game to be a swarm. Meanwhile, starling murmurations that are formed as a result of each bird trying to escape a predator might not be considered a swarm since they aren't intentionally working together to create those shapes (i.e., no cooperation towards a group objective).

From Table~\ref{tbl:Swarm_defs}, we find some of the most common factors that must exist for a system in question to be a ``swarm" include quantity, some notion/appearance of cooperation or a shared objective, decentralization control, and self-organization:

\begin{itemize}
    \item \textbf{Multiple agents}: a system consisting of more than one agent ($N > 1$).\\

    \item \textbf{Similar Agents}: agents share the same capabilities/dynamics ($f_i(p,u,w) = f_k(p,u,w)$~for all  $i,k \in \{1,...N\}$).\\

    \item \textbf{Recognizable group behavior}: the agents’ local actions and interactions must produce a recognizable group-level behavior to an observer ($B^\indexx = 1$ if $M^\indexx \in \eta^\indexx$). Specifically, the set~$\eta^\indexx$ must be defined with the dimension of~$\SS_j$ in~\eqref{eq:trajectories_behavior} being strictly smaller than the dimension of~$\PP^N$. That is, the recognized group behavior imposes a constraint of non-zero measure on the state space of the system.  \\ 

    \item \textbf{Agency}: each agent must have a basic level of agency in that it must sense/perceive the environment and take a local control action. Under our framework, the internal perceptions of agents within the system are latent variables or constructs. For our purposes agency then means an agent is able to modify its own input~$u_i$ based on the states of others or its own latent/hidden states~$u_i = g_i(P, z_i, t)$. \\

    \item \textbf{Local interactions}: agents should interact with one another, influencing their own and others' behaviors (i.e., $\dot{p}_i = f_i(P, u_i, t)$ ). \\

    \item \textbf{Decentralized with no leader}: the system should be decentralized with no designated leaders or external control. We acknowledge this one is still semantically complicated and discuss further in Remark~\ref{re:leaders}.
\end{itemize} 

We apply our objective definitions presented above to the works in Table~\ref{tbl:Swarm_defs} and summarize the results in Table~\ref{tbl:swarm_conditions}. We notice that most works do fit our proposed checklist but a few are missing some features that others might consider important in defining a swarm.

\begin{table}[htbp]
\centering
\caption{Comparing proposed necessary conditions of a swarm to previous works and applying the conditions to various examples.}
\footnotesize
\begin{tabular}{|>{\raggedright\arraybackslash}p{3.4cm}|>{\raggedright\arraybackslash}p{1cm}|>{\raggedright\arraybackslash}p{1cm}|>{\raggedright\arraybackslash}p{1.7cm}|>{\raggedright\arraybackslash}p{1cm}|>{\raggedright\arraybackslash}p{1.6cm}|>{\raggedright\arraybackslash}p{1.7cm}|
>{\raggedright\arraybackslash}p{1.5cm}|}
\hline
\rowcolor{gray!30}
\textbf{Source} & \textbf{Multiple Agents} &  \textbf{Similar Agents} &  \textbf{Recognizable Group Behavior} &  \textbf{Agency} &  \textbf{Local Interactions} &  \textbf{Decentralized with No Leader} & \textbf{Meets Swarm Criteria?}\\
\hline

Rubenstein et al. \cite{MR-CA-RN:12}
& \checkmark & & \checkmark & \checkmark & \checkmark & \checkmark & \\

\hline

 Hamann \cite{HH:18} & \checkmark & \checkmark & \checkmark & \checkmark & \checkmark & \checkmark  & \checkmark\\

\hline

Haider \& Schmidt~\cite{AH-AS:22} & \checkmark & & \checkmark & \checkmark & \checkmark & & \\

\hline

Ligot \& Birattari \cite{AL-MB:18} & \checkmark & \checkmark & \checkmark & \checkmark & \checkmark & \checkmark & \checkmark \\

\hline 

Brambilla et al.~\cite{MB-EF-MB-MD:13} & \checkmark & & \checkmark & \checkmark & \checkmark & \checkmark &\\

\hline 

Salman et al. \cite{MS-DGR-MB:24} & \checkmark & & \checkmark & \checkmark & \checkmark & & \\

\hline 

Kolling et al. \cite{AK-PW-NC-KS-ML:15} & \checkmark & & \checkmark & \checkmark & \checkmark & & \\

\hline 

Birattari et al. \cite{MB-AL-DB-MB-GF-LG-DGR-KH-MK-JK:19} & \checkmark & & \checkmark & \checkmark & \checkmark & \checkmark & \\

\hline

Arnold et al. \cite{RA-KC-BA-CK:19} & \checkmark & & \checkmark & \checkmark & \checkmark & & \\ 

\hline

Navarro \& Mattia~\cite{IN-FM:13} & \checkmark & \checkmark & \checkmark & \checkmark & \checkmark & \checkmark & \checkmark\\

\hline

Beni \& Wang~\cite{GB-JW:93} & \checkmark & \checkmark & \checkmark & \checkmark & \checkmark & \checkmark & \checkmark\\

\hline
Calderon et al~\cite{CCA-JCBT-RFSO:22} & \checkmark & \checkmark & \checkmark & \checkmark & \checkmark & &\\

\hline

Abbass \& Mostaghim \cite{HA-SM:25} &  \checkmark & & \checkmark & \checkmark & \checkmark & &\\

\hline
\rowcolor{gray!30} \textbf{Examples from Fig. \ref{fig:panel}} & & & & & & &\\
\hline
Ducky Derby Race
& \checkmark & \checkmark & \checkmark &  & \checkmark &  & \\

\hline

Drone Show 
~

& \checkmark & \checkmark & \checkmark &  & \checkmark &  & \\

\hline

Starling Murmuration 
& \checkmark & \checkmark & \checkmark & \checkmark & \checkmark & \checkmark & \checkmark\\

\hline

\hline
\rowcolor{gray!30} \textbf{Examples from Fig. \ref{fig:perceptions}} & & & & & & &\\
\hline

a) No emergence
& & &  &  &  &  & \\

\hline

b) No feedback 
& \checkmark & \checkmark & \checkmark &  &  &  & \\

\hline

c) Simple feedback
& \checkmark & \checkmark & \checkmark & \checkmark & \checkmark & \checkmark & \checkmark\\

\hline

d) Distributed control
& \checkmark & \checkmark & \checkmark & \checkmark & \checkmark &  & \\

\hline

\hline
\end{tabular}
\label{tbl:swarm_conditions}
\end{table}

From this non-exhaustive list of sources, only a few explicitly note the necessity for agents to be similar. Other works either don't specify this condition or allow for heterogeneous swarms~\cite{MR-CA-RN:12, AH-AS:22, MB-EF-MB-MD:13, MS-DGR-MB:24, AK-PW-NC-KS-ML:15, MB-AL-DB-MB-GF-LG-DGR-KH-MK-JK:19, RA-KC-BA-CK:19, HA-SM:25}. Although there are various works on heterogeneous swarms, we propose that these should not be considered to be a single swarm but a group of multiple swarms of various agents. Some works~\cite{AH-AS:22, MS-DGR-MB:24, AK-PW-NC-KS-ML:15, CCA-JCBT-RFSO:22} allow for the agents in the system to be aware of the group's goal and work cooperatively towards (i.e. distributed control) which therefore fails to meet the last condition. This will be discussed further as an exciting avenue for future research of swarms in Section~\ref{se:interactions}. 
 
The components of this list of necessary conditions isn't exactly novel as many of the definitions in Table~\ref{tbl:Swarm_defs} incorporate these ideas (some more than others) as can be see in Table~\ref{tbl:swarm_conditions}.

Table~\ref{tbl:swarm_conditions} will now be used to evaluate our motivating examples from Fig.~\ref{fig:panel} and the running milling example. 

\begin{example}[Are the milling agents a swarm?]
{\rm
Let us now revisit Example~\ref{ex:intro} again and answer Problem~\ref{pr:swarm} with respect to each of the four different contexts provided in Example~\ref{ex:emergence} and shown in Fig.~\ref{fig:perceptions}. By simply applying our swarm checklist we can conclude (summarized in Table~\ref{tbl:swarm_conditions}):

\noindent Example (a) is \textbf{not a swarm} because it is one singular entity therefore it fails all the conditions. 

\noindent Example (b) is \textbf{not a swarm} because the agents cannot take action based on its perception let alone interact with one another. 

\noindent Example (c) \textbf{is a swarm} because it satisfies all the conditions. 

\noindent Example (d) is \textbf{not a swarm} because the agents are aware of the actions they need to take to create the global behavior (i.e. the circle formation).
}
\end{example}

Although the knowledge of whether or not the conditions are satisfied are still dependent on what the observer knows about the system, these conditions give us a way to formally present whether something is a swarm by ensuring the information needed to make that decision is clearly stated.

There are a very large number of different ways to form circles, based on different applications, sensor configurations, and feedback mechanisms~\cite{HL-ZYH-AH:21}. Regardless of the final appearance, and even resolution and scope of an observer, the group behavior can be the same; however, the type of emergence that occurs is not fixed and dependent on an observer's perception and knowledge. On the contrary, we find that it is \textit{entirely dependent} on it. 

\begin{proposition}[Equifinality of Emergent Processes]\label{pr:emergent}
For any context~$\context$ in which a system~$P(t)$ exhibits an observable behavior~$B$ via emergence of type~$T \in \{\text{I, II, III} \}$, there exists a different context~$\mathbb{C}'$ in which the same observable information and behavior~$B$ emerges via type~$T-1$ emergence.
\end{proposition}

Proposition~\ref{pr:emergent} reveals something interesting and even potentially controversial among some researchers studying emergence. As is emphasized in \cite{CG:25, BRJ:10, PWA:72}, if emergence is defined as the ``surprise" of the observer, then emergence is level of emergence changes as we increase our knowledge/understanding of the system. According to Gershenson, ``so then emergence would be a measure of our ignorance, and then it would be reduced once we understood the mechanisms behind emergent properties''~\cite{CG:25}. We tend to agree that it does not make sense that these classifications can change with our knowledge base, but rather they must be defined according to a particular knowledge base. So indeed based on today's understanding, perhaps the emergence of consciousness and emotions like love and hate are ``strongly emergent" (i.e., not understood) results of interacting neurons, but one day when we have a universal simulator that can exhibit these properties, by definition we will have turned it into a weakly emergent process.

Returning now to the three images in Fig.~\ref{fig:panel}, we can easily see that all of them can be viewed as a system with many components to produce a similar (if not the exact same) group behavior~$B_\indexx$. In Fig. \ref{fig:panel}(a) we see a system that is intuitively not a swarm. In~(c) we see a self-organized swarm of starlings which is close to how the term originated. In (b) depending on who you ask you might get a different answer. In particular laymen and engineers tend to answer yes while biologists/psychologists tend to answer no. Who is right?

\section{Future Outlook for Swarm Robotics Research}\label{se:outlook}

There is still a lot of work left to do before we can truly understand how to design a large team or swarm of robots to effectively solve real-world problems. And while there are many exciting and promising algorithms and systems showing the potential power of swarms, we are still lacking general design principles for creating them. How can we design swarms that enjoy the precision and reliability of traditionally engineered systems while also having the adaptability and resilience of self-organizing biological systems? 

From Conway's Game of life in 1970~\cite{LSS-PES:78, VB:06} to Wolfram’s work on characterizing the behaviors of simple cellular automata \cite{SW:02}, Agent-Based Modeling (ABM) and Simulation of Complex Systems in virtual environments has long demonstrated the ability to generate amazingly complex behaviors from a very simple set of base rules. In these communities, the emergent behaviors are generally ``discovered" by playing with the different local-interaction or environmental rules. Once discovered, they can better understood in terms of the exact conditions that give rise to them. Unfortunately, it is extremely difficult to transfer any of these virtually discovered behaviors to physical deployments due to the inherently chaotic nature of these systems. 

Engineers working on swarms generally tend to do the opposite. Instead of ``discovering" the emergent behaviors, they first decide what specific emergent behavior they want and then design the local-interaction rules to produce the intended behavior. This generally means these systems can only be deployed in controlled, nonstationary environments.

We believe bridging the ideas in these two different research fields is precisely the way forward in deploying swarming robots in the wild. As not to be redundant with many other great surveys and review articles discussing the current state of the art and future research directions~\cite{SC-AAP-PD-SS-VK:18, LY-JY-SY-BW-BJN-LZ:21, LI-DN-MS:01, AK-PW-NC-KS-ML:15, MS-MU-MS-WE:20, MD-GT-VT:21, ARC-SS-KG:22, MB-EF-MB-MD:13, YM-SGP:09, JCB-YAK:13, MMS-ZS-AA-HM-MHY-NKB-FH:23}, we highlight some paths forward that may help in bridging these two fields. 

\subsection{Morphogenetic Engineering}
We begin by highlighting a newer field of research launched in 2009~\cite{RD-HS-OM:13, YJ-YM:11, RD-HS-OM:12}. This field is concerned with combining the rigid architectures of artificial/engineered systems with the self-organizing properties of natural/biological systems. We believe the goals of this field are very well aligned with the ideas presented in this proposal, and in particular their field relies heavily on the concept of emergent properties and focuses on the \textit{process} on which things emerge. They categorize different systems in terms of `Constructing', `Coalescing', `Developing', or `Generating'. We don't enter into the details here but note that there are many novel and interesting systems that have been developing within Morphogenetic Engineering and provides a way of distinguishing a few of the different processes in which the emergent effects are created. However, even with successes like the remarkable Strandbeest~\cite{TJ:08, SP-15}, general design principles are still lacking and actually building such systems is a very human-intensive and non-systematic process.

\subsection{Swarm Analytics and Swarm Chemistry}
Turning now to more mathematical modeling and analysis techniques, we highlight the recent ideas of swarm analytics from the engineering and control domain alongside swarm chemistry from the complex systems, artificial life, and ABM domain. 

We hope this paper already demonstrates the values of swarm analytics as we have leveraged it in enabling an objective discussion of the enigmatic concepts of swarming or emergence~\cite{AJH-AH-DJR-HAA:23, AJH-ASMH-DJR-HAA:24, JH-MB:21}. We believe this has been a necessary and missing step towards the formalization and development of generalized design principles and tools for the study of swarms. 

On the other hand, Swarm Chemistry (originally developed by Sayama in \cite{HS:09} is a concrete step towards bridging the emergent behaviors discovered through simulation to real robots through a both visual and analytical understanding of the conditions that give rise to different behaviors. This can enable engineers to make use of these behaviors and know exactly how to ``control" or ``manage" emergent behaviors by changing the conditions around the agents. While Swarm Chemistry was originally developed in the context of heterogeneous swarms to understand how two different ``species" of swarms might interact with each other and produce their own second-order emergent effects, new work attempts to further establish this connection to chemistry even for homogeneous agents in seek of novel concepts like a Swarm Temperature or Pressure~\cite{RV-CM-DSB-CN:23, HH-KJ-JL:21} similar to how we understand what environmental conditions to put $H_2 0$ molecules in if we want the emergent property of ice versus liquid versus gas. 

Within this we highlight also research pushing more the automated ``discovery" of such emergent behaviors using computational methods~\cite{RV-KZ-CM-DSB-CN:24, CM-DB:23, CM-VR-RV-CN-DSD-DSB:25, JG-ALC:18, DB-RT-OH-SL:18, JG-PU-ALC:13}. 
Today this is a human-intensive process that requires domain experts to design objective functions or implement simulations and understand them in various complex systems applications. This line of research aims to make this process more systematic, efficient, and automated.

\subsection{Agent-to-Swarm and Swarm-to-Swarm Interactions}\label{se:interactions}
As a concrete step towards how to design real-world deployable swarms, we believe the most promising works are ones that already combine some sort of top-down guidance with bottom-up emergence similar to a sheepdog guiding a herd of sheep~\cite{PK-SH-HV:02dynamics, PK-SH-HV:02, TN-TN:06, SR-QM-SHY:11, BB-MT:12}. 

\begin{remark}[Heterogeneous Swarms?]\label{re:leaders}
{\rm
In this paper we have made a deliberate choice to include the requirement of ``similar" (e.g., at least idiosyncratic, if not homogeneous) when defining a single swarm. This means under this framework the concept of a ``heterogeneous swarm" is not valid. We believe this to simply be a semantic choice at this point but helpful in not entering into philosophical debates over whether two interacting swarms that produce a new emergent behavior should be called its own swarm. 
}
\end{remark}

Regardless of what we call it, the main ideas here are that more research is needed in understanding how a homogeneous swarm can be controlled or managed both externally and internally. While there are indeed many systems that demonstrate this combination, we again seek more generalized methods of designing both sides of these systems (the ``sheep" swarm to be manipulated along with the ``dog" that can manage the swarm). Pinning control is also a broad area in which the similar idea of a single (or few) ``leader" agents are controlling a swarm from within~\cite{FC-ZC-LX-ZXL-ZY:09, XFW-GC:02, XW-XL-JL:24}. It is also worth mentioning that the controlling/orchestrating agent may indeed be a human too~\cite{JW-LM:07, AK-PW-NC-KS-ML:15, AH-HAA:14, PW-SN-ML-AK-NC-KS:12}. 

Finally, we would like to highlight an even broader class of ``Ultra-Large Scale Systems" and control that this paper fits within. In these systems the managing agents are necessarily a part of the system that needs managing, and similar issues and concepts of ``emergence" will be highly relevant in both fields~~\cite{LP-PB-WPMHM:25}.

\section{Conclusions}\label{se:conclusions}
This paper has presented a deep dive into the terms ``emergence" and ``swarm", distinguishing the two as separate concepts while showing exactly how they are related. The definitions and framework presented here do not disagree with most related studies on emergence or swarms in the literature. Instead, it exemplifies the role of the generally implicit or invisible observer. Definitions of scope, resolution, context, and the separation between a system and its environment are generally implicit in the literature, and often times not even fixed. 

This work aims to provide a unifying framework capturing all aspects of emergence and swarms to be comparable in an objective way. By explicitly describing the role that both the knowledge and vantage point of an observer plays in relation to understanding the process in which an emergent property ``emerges'', we hope to help not only new researchers in the field of complex systems and swarm robotics to taxonomically separate different forms and types of emergence, but also enable active researchers to objectively present their methods and results with less risk of misunderstanding. Commensurable and objective discussions about emergence and swarms can only occur after both the vantage point \textit{and} tacit knowledge of an observer about a context~$\mathbb{C}$ are explicitly defined.

Separately from emergence, this has \textbf{not} been a comprehensive survey of swarms or even swarming robots for which many great surveys and review articles already exist. The potential benefits of swarm systems (robustness, scalability, flexibility, expandability)\cite{YT-ZZ:13, JCB-YAK:13, BK-FC:15, SC-AAP-PD-SS-VK:18} has been explored in many different application areas including autonomous navigation \cite{AMS-MK:16, AD-MK-SR:17, HJML-DAL:22, AO-MG-AK-MDH-RG:19, QW-HZ:21, DAL-GMBO:17, RZ-VK-EW-JP-SB:15, ASK-GM-RRB-MSC:17, RM-CSJ-CV:18, JY-RX-XX-YS:20, VG-AS-RT:22, BY-YD-YJ-KH:15, OYM:21, ZZ-YT:13, JPM-CN:23, JL-YT:14, ZZ-JL-JL-YT:14, QT-FY-LD:16, JL-YT:16}, spatial organization \cite{ZN-QZ-XW-FW-TH:23,DS-CP-GB:18, FB-MG-RN:21, MG-JC-TJD-RG:14, MG-JC-WL-TJD-RG:14, VG:05, DS-LV:19}, and collective decision-making \cite{JW-CM:03, SG-JG-MA-CJ-GT:09, AC-SG-OD-MZ-MD:11, AG-AC-FM-LM-MD:10, CACP-HZ:11, AA-MA-JC-ET:21}.

While the above works paint a much broader picture of everything related to swarms including applications, we have focused more narrowly on the intersection between systems and control engineering methods and more scientific approaches to demonstrate the value of each side. To truly harness the power of swarms in engineered systems, we must bridge this gap. Beyond the superficial motivating connection of ``bio-inspired robotics" before continuing to work in isolation with engineers/roboticists and no biologists, we must actually collaborate on the actual ideas, methods, and processes too. It's easy enough to call a robotic system bio-inspired because it looks like something biological, but we hope this paper has shown precisely why this means nothing about the inner workings of how it was produced. This means that while they can look similar in controlled environments, they reveal vastly different properties when perturbed. By better understanding not only how these biological processes are created but more importantly the methods of studying biological processes, swarm engineers can gain value by allowing their engineered systems to sometimes exhibit surprises. On the other hand, agent-based methods must be applied to physical systems in order for the emergent behaviors found to be practically useful.

\section*{Acknowledgments}

This material is based upon research supported by the U. S. Office of Naval Research under award number N00014-22-1-2207.

\bibliographystyle{ieeetr}
\bibliography{survey}

\begin{thebibliography}{100}

\bibitem{TSK:97}
T.~S. Kuhn, {\em The structure of scientific revolutions}.
\newblock University of Chicago Press, 1997.

\bibitem{LB:1877}
L.~Boltzmann, ``On the relationship between the second fundamental theorem of the mechanical theory of heat and probability calculations regarding the conditions for thermal equilibrium,'' {\em Sitzungsberichte der Kaiserlichen Akademie der Wissenschaften, Mathematisch-Naturwissenschaftliche Classe, Abteilung II}, vol.~76, pp.~373--435, 1877.

\bibitem{ETJ:57}
E.~T. Jaynes, ``Information theory and statistical mechanics,'' {\em Physical review}, vol.~106, no.~4, p.~620, 1957.

\bibitem{HBC-HLS:98}
H.~B. Callen and H.~L. Scott, ``Thermodynamics and an introduction to thermostatistics,'' 1998.

\bibitem{PWA:72}
P.~W. Anderson, ``More is different: Broken symmetry and the nature of the hierarchical structure of science.,'' {\em Science}, vol.~177, no.~4047, pp.~393--396, 1972.

\bibitem{SAK:92}
S.~A. Kauffman, ``Origins of order in evolution: self-organization and selection,'' in {\em Understanding origins: Contemporary views on the origin of life, mind and society}, pp.~153--181, Springer, 1992.

\bibitem{NG-CW:11}
N.~Goldenfeld and C.~Woese, ``Life is physics: evolution as a collective phenomenon far from equilibrium,'' {\em Annu. Rev. Condens. Matter Phys.}, vol.~2, no.~1, pp.~375--399, 2011.

\bibitem{RBL-DP-JS=BPS-PW:00}
R.~B. Laughlin, D.~Pines, J.~Schmalian, B.~P. Stojkovi{\'c}, and P.~Wolynes, ``The middle way,'' {\em Proceedings of the National Academy of Sciences}, vol.~97, no.~1, pp.~32--37, 2000.

\bibitem{DW:12}
D.~Wallace, {\em The emergent multiverse: Quantum theory according to the Everett interpretation}.
\newblock OUP Oxford, 2012.

\bibitem{GFRE:06}
G.~F.~R. Ellis, ``Physics and the real world,'' {\em Foundations of Physics}, vol.~36, pp.~227--262, 2006.

\bibitem{IN:1687}
I.~Newton, {\em Philosophi{\ae} Naturalis Principia Mathematica}.
\newblock Londini: Jussu Societatis Regi{\ae} ac Typis Josephi Streater, 1687.

\bibitem{PD-LFA:05}
P.~Dayan and L.~F. Abbott, {\em Theoretical neuroscience: computational and mathematical modeling of neural systems}.
\newblock MIT press, 2005.

\bibitem{JHH:95}
J.~H. Holland, ``Hidden order,'' {\em Business Week-Domestic Edition}, vol.~21, pp.~93--136, 1995.

\bibitem{OS:16}
O.~Sporns, {\em Networks of the Brain}.
\newblock MIT press, 2016.

\bibitem{SAL:98}
S.~A. Levin, ``Ecosystems and the biosphere as complex adaptive systems,'' {\em Ecosystems}, vol.~1, pp.~431--436, 1998.

\bibitem{DJC:97}
D.~J. Chalmers, {\em The conscious mind: In search of a fundamental theory}.
\newblock Oxford Paperbacks, 1997.

\bibitem{JRS:92}
J.~R. Searle, {\em The rediscovery of the mind}.
\newblock MIT press, 1992.

\bibitem{JK:00}
J.~Kim, ``Mind in a physical world: An essay on the mind-body problem and mental causation,'' MIT press, 2000.

\bibitem{CGL:90}
C.~G. Langton, ``Computation at the edge of chaos: Phase transitions and emergent computation,'' {\em Physica D: nonlinear phenomena}, vol.~42, no.~1-3, pp.~12--37, 1990.

\bibitem{MM:09}
M.~Mitchell, {\em Complexity: A guided tour}.
\newblock Oxford university press, 2009.

\bibitem{EMI:07}
E.~M. Izhikevich, {\em Dynamical systems in neuroscience}.
\newblock MIT press, 2007.

\bibitem{HRM-FJV:12}
H.~R. Maturana and F.~J. Varela, {\em Autopoiesis and cognition: The realization of the living}, vol.~42.
\newblock Springer Science \& Business Media, 2012.

\bibitem{GT:04}
G.~Tononi, ``An information integration theory of consciousness,'' {\em BMC neuroscience}, vol.~5, pp.~1--22, 2004.

\bibitem{TWD:11}
T.~W. Deacon, {\em Incomplete nature: How mind emerged from matter}.
\newblock WW Norton \& Company, 2011.

\bibitem{FJV-ET-ER:17}
F.~J. Varela, E.~Thompson, and E.~Rosch, {\em The embodied mind, revised edition: Cognitive science and human experience}.
\newblock MIT press, 2017.

\bibitem{HK-AB-FB-BF-OH-SI:16}
H.~Kopetz, A.~Bondavalli, F.~Brancati, B.~Fr{\"o}mel, O.~H{\"o}ftberger, and S.~Iacob, ``Emergence in cyber-physical systems-of-systems (cpsoss),'' {\em Cyber-Physical Systems of Systems: Foundations--A Conceptual Model and Some Derivations: The AMADEOS Legacy}, pp.~73--96, 2016.

\bibitem{VD:94}
V.~Darley, ``Emergent phenomena and complexity,'' {\em Artificial Life}, vol.~4, pp.~411--416, 1994.

\bibitem{SF-MF-HWC:13}
S.~Ferreira, M.~Faezipour, and H.~W. Corley, ``Defining and addressing the risk of undesirable emergent properties,'' in {\em 2013 IEEE International Systems Conference (SysCon)}, pp.~836--840, IEEE, 2013.

\bibitem{RAH-NOS-GM-ES:23}
R.~A. Haugen, N.~O. Skeie, G.~Muller, and E.~Syverud, ``Detecting emergence in engineered systems: A literature review and synthesis approach,'' {\em Systems Engineering}, vol.~26, no.~4, pp.~463--481, 2023.

\bibitem{BY-JZ-AL-JW-ZW-MY-KL-MM-PC:24}
B.~Yuan, J.~Zhang, A.~Lyu, J.~Wu, Z.~Wang, M.~Yang, K.~Liu, M.~Mou, and P.~Cui, ``Emergence and causality in complex systems: A survey of causal emergence and related quantitative studies,'' {\em Entropy}, vol.~26, no.~2, p.~108, 2024.

\bibitem{QL-JW-QX-YH-HZ:15}
Q.~Li, J.~Wang, Q.~Xu, Y.~Huang, and H.~Zhu, ``A formal framework for reasoning emergent behaviors in swarm robotic systems,'' in {\em 2015 20th International Conference on Engineering of Complex Computer Systems (ICECCS)}, pp.~150--159, IEEE, 2015.

\bibitem{JD-YD-LM:06}
J.~Deguet, Y.~Demazeau, and L.~Magnin, ``Elements about the emergence issue: A survey of emergence definitions,'' {\em Complexus}, vol.~3, no.~1-3, pp.~24--31, 2006.

\bibitem{RC-CC:96}
R.~Conte and C.~Castelfranchi, ``Simulating multi-agent interdependencies. a two-way approach to the micro-macro link,'' in {\em Social science microsimulation}, pp.~394--415, Springer, 1996.

\bibitem{BBL-BM:11}
B.~B. Lichtenstein and B.~McKelvey, ``Four types of emergence: a typology of complexity and its implications for a science of management,'' {\em International Journal of Complexity in Leadership and Management}, vol.~1, no.~4, pp.~339--378, 2011.

\bibitem{BM:92}
B.~McLaughlin, ``The rise and fall of british emergentism,'' {\em Emergence or reduction}, pp.~49--93, 1992.

\bibitem{MAB-PH:08}
M.~A. Bedau and P.~Humphreys, {\em Emergence: Contemporary readings in philosophy and science}.
\newblock MIT press, 2008.

\bibitem{MP-FB-AJR:09}
M.~Prokopenko, F.~Boschetti, and A.~J. Ryan, ``An information-theoretic primer on complexity, self-organization, and emergence,'' {\em Complexity}, vol.~15, no.~1, pp.~11--28, 2009.

\bibitem{MSA:95}
M.~Archer, {\em Realist social theory: The morphogenetic approach}.
\newblock Cambridge university press, 1995.

\bibitem{JSC:84}
J.~S. Coleman, ``Micro foundations and macrosocial behavior,'' {\em Angewandte Sozialforschung anc AIAS Informationen Wien}, vol.~12, no.~1-2, pp.~25--37, 1984.

\bibitem{WR-TV:17}
W.~Raub and T.~Voss, ``Micro-macro models in sociology: antecedents of coleman’s diagram,'' {\em Social dilemmas, institutions, and the evolution of cooperation}, pp.~11--36, 2017.

\bibitem{GM:05}
G.~Manzo, ``Variables, mechanisms, and simulations: Can the three methods be synthesized?,'' {\em Revue fran{\c{c}}aise de sociologie}, vol.~46, no.~1, pp.~37--74, 2005.

\bibitem{RJ-JWM:11}
R.~Jepperson and J.~W. Meyer, ``Multiple levels of analysis and the limitations of methodological individualisms,'' {\em Sociological Theory}, vol.~29, no.~1, pp.~54--73, 2011.

\bibitem{PA-TF-NF:08}
P.~Abell, T.~Felin, and N.~Foss, ``Building micro-foundations for the routines, capabilities, and performance links,'' {\em Managerial and decision economics}, vol.~29, no.~6, pp.~489--502, 2008.

\bibitem{DJC:06}
D.~Chalmers, ``Strong and weak emergence,'' {\em The re-emergence of emergence}, vol.~675, pp.~244--256, 2006.

\bibitem{RA:04}
R.~Abbott, ``Emergence, entities, entropy, and binding forces,'' in {\em The Agent 2004 Conference on: Social Dynamics: Interaction, Reflexivity, and Emergence}, p.~17, 2004.

\bibitem{EB:02}
E.Bonabeau, ``Predicting the unpredictable,'' {\em Harvard Business Review}, vol.~80, no.~3, pp.~109--134, 2002.

\bibitem{EMAR-MS-MSC:99}
E.~M.~A. Ronald, M.~Sipper, and M.~S. Capcarr{\`e}re, ``Design, observation, surprise! a test of emergence,'' {\em Artificial Life}, vol.~5, no.~3, pp.~225--239, 1999.

\bibitem{JLC:94}
J.~L. Casti, ``Complexification: Explaining a paradoxical world through the science of surprise,'' 1994.

\bibitem{SMC-AP:24}
S.~M. Carroll and A.~Parola, ``What emergence can possibly mean,'' {\em arXiv preprint arXiv:2410.15468}, 2024.

\bibitem{SJ:02}
S.~Johnson, ``Emergence: The connected lives of ants, brains, cities, and software,'' 2002.

\bibitem{MAB:97}
M.~A. Bedau, ``Weak emergence,'' {\em Philosophical perspectives}, vol.~11, pp.~375--399, 1997.

\bibitem{MM-CMS:11}
M.~Mnif and C.~M{\"u}ller-Schloer, ``Quantitative emergence,'' {\em Organic Computing—A Paradigm Shift for Complex Systems}, pp.~39--52, 2011.

\bibitem{SM-LR:15}
S.~Mittal and L.~Rainey, ``Harnessing emergence: The control and design of emergent behavior in system of systems engineering,'' in {\em Proceedings of the conference on summer computer simulation}, pp.~1--10, 2015.

\bibitem{CG:25}
C.~Gershenson, ``Self-organizing systems: what, how, and why?,'' {\em npj Complexity}, vol.~2, no.~1, p.~10, 2025.

\bibitem{YBY:04}
Y.~Bar-Yam, ``A mathematical theory of strong emergence using multiscale variety,'' {\em Complexity}, vol.~9, no.~6, pp.~15--24, 2004.

\bibitem{EB-JLD:11}
E.~Bonabeau and J.~Dessalles, ``Detection and emergence,'' {\em arXiv preprint arXiv:1108.4279}, 2011.

\bibitem{JPC:94}
J.~P. Crutchfield, ``The calculi of emergence: computation, dynamics and induction,'' {\em Physica D: Nonlinear Phenomena}, vol.~75, no.~1-3, pp.~11--54, 1994.

\bibitem{AJR:07}
A.~J. Ryan, ``Emergence is coupled to scope, not level,'' {\em Complexity}, vol.~13, no.~2, pp.~67--77, 2007.

\bibitem{GSB-CG-NF:17}
G.~Santamar{\'\i}a-Bonfil, C.~Gershenson, and N.~Fern{\'a}ndez, ``A package for measuring emergence, self-organization, and complexity based on shannon entropy,'' {\em Frontiers in Robotics and AI}, vol.~4, p.~10, 2017.

\bibitem{AKS:08}
A.~K. Seth, ``Measuring emergence via nonlinear granger causality.,'' in {\em alife}, vol.~2008, pp.~545--552, 2008.

\bibitem{MB:02}
M.~Bedau, ``Downward causation and the autonomy of weak emergence,'' {\em Principia: an international journal of epistemology}, vol.~6, no.~1, pp.~5--50, 2002.

\bibitem{RH:85}
R.~Harr{\'e}, ``The philosophies of science,'' 1985.

\bibitem{NAB:94}
N.~A. Baas, ``Emergence, hierarchies, and hyperstructures,'' {\em Artificial Life III, SFI Studies in the Science of Complexity, XVII}, pp.~515--537, 1994.

\bibitem{JK:92}
J.~Kim, ``‘downward causation’in emergentism and nonreductive physicalism,'' {\em Emergence or reduction}, pp.~119--138, 1992.

\bibitem{TO:20}
T.~O’connor, ``Emergent properties,'' 2020.

\bibitem{NL:95}
N.~Luhmann, {\em Social systems}.
\newblock Stanford university Press, 1995.

\bibitem{JHH:00}
J.~Holland, {\em Emergence: From chaos to order}.
\newblock OUP Oxford, 2000.

\bibitem{PK-SH-HV:02dynamics}
P.~Kachroo, S.~A. Shedied, and H.~Vanlandingham, ``Dynamic programming solution for a class of pursuit evasion problems: the herding problem,'' {\em IEEE Transactions on Systems, Man, and Cybernetics, Part C (Applications and Reviews)}, vol.~31, no.~1, pp.~35--41, 2002.

\bibitem{PK-SH-HV:02}
P.~Kachroo, S.~A. Shedied, and H.~Vanlandingham, ``Pursuit evasion: The herding noncooperative dynamic game—the stochastic model,'' {\em IEEE Transactions on Systems, Man, and Cybernetics, Part C: Applications and Reviews}, vol.~32, no.~1, pp.~37--42, 2002.

\bibitem{TN-TN:06}
T.~Miki and T.~Nakamura, ``An effective simple shepherding algorithm suitable for implementation to a multi-mobile robot system,'' in {\em First International Conference on Innovative Computing, Information and Control (ICICIC'06)}, vol.~3, pp.~161--165, IEEE, 2006.

\bibitem{SR-QM-SHY:11}
S.~Razali, Q.~Meng, and S.~Yang, ``Immune inspired cooperative mechanism with refined low-level behaviors for multi-robot shepherding,'' {\em International Journal of Computational Intelligence and Applications}, vol.~10, no.~1, pp.~1--18, 2011.

\bibitem{BB-MT:12}
B.~Bennett and M.~Trafankowski, ``A comparative investigation of herding algorithms,'' in {\em Proceedings of the Symposium on Understanding and Modelling Collective Phenomena (UMoCoP)}, (Birmingham, UK), pp.~33--38, 2012.

\bibitem{AJH-AH-DJR-HAA:23}
A.~J. Hepworth, A.~Hussein, D.~J. Reid, and H.~A. Abbass, ``Swarm analytics: Designing information markers to characterise swarm systems in shepherding contexts,'' {\em Adaptive behavior}, vol.~31, no.~4, pp.~323--349, 2023.

\bibitem{FB-JC-SM:09}
F.~Bullo, J.~Cort{\'e}s, and S.~Martinez, {\em Distributed control of robotic networks: a mathematical approach to motion coordination algorithms}.
\newblock Princeton University Press, 2009.

\bibitem{JM-MEB-BAF:04}
J.~Marshall, M.~Broucke, and B.~Francis, ``Formations of vehicles in cyclic pursuit,'' {\em IEEE Transactions on Automatic Control}, vol.~49, no.~11, pp.~1963--1974, 2004.

\bibitem{ZL-MEB-BAF:04}
Z.~Lin, M.~Broucke, and B.~Francis, ``Local control strategies for groups of mobile autonomous agents,'' {\em IEEE Transactions on Automatic Control}, vol.~49, no.~4, pp.~622--629, 2004.

\bibitem{QW-YW-HZ:16}
Q.~Wang, Y.~Wang, and H.~Zhang, ``The formation control of multi-agent systems on a circle,'' {\em International Journal of Control, Automation and Systems}, vol.~14, no.~5, pp.~1234--1242, 2016.

\bibitem{CW-GX:17}
C.~Wang and G.~Xie, ``Limit-cycle-based decoupled design of circle formation control with collision avoidance for anonymous agents in a plane,'' {\em IEEE Transactions on Automatic Control}, vol.~62, no.~12, pp.~6560--6567, 2017.

\bibitem{RZ-YL-DS:15}
R.~Zheng, Y.~Liu, and D.~Sun, ``Enclosing a target by nonholonomic mobile robots with bearing-only measurements,'' {\em Automatica}, vol.~53, pp.~400--407, 2015.

\bibitem{XY-LL:17}
X.~Yu and L.~Liu, ``Moving target circular formation control of multiple non-holonomic vehicles without global position measurements,'' {\em IEEE Transactions on Automatic Control}, vol.~62, no.~7, pp.~3448--3454, 2017.

\bibitem{NM-NM-AJ-KD:09}
N.~Moshtagh, N.~Michael, A.~Jadbabaie, and K.~Daniilidis, ``Vision-based, distributed control laws for motion coordination of nonholonomic robots,'' {\em IEEE Transactions on Robotics}, vol.~25, no.~4, pp.~851--860, 2009.

\bibitem{NC-MDM-AG-AG:08}
N.~Ceccarelli, M.~D. Marco, A.~Garulli, and A.~Giannitrapani, ``Collective circular motion of multi-vehicle systems,'' {\em Automatica}, vol.~44, no.~12, pp.~3025--3035, 2008.

\bibitem{YH-RA:08}
Y.~Hou and R.~Allen, ``Behaviour-based circle formation control simulation for cooperative uuvs,'' in {\em Proceedings of the 17th IFAC World Congress}, vol.~41, pp.~119--124, Elsevier, 2008.

\bibitem{RZ-ZL-MF-DS:15}
R.~Zheng, Z.~Lin, M.~Fu, and D.~Sun, ``Distributed control for uniform circumnavigation of ring-coupled unicycles,'' {\em Automatica}, vol.~53, pp.~23--29, 2015.

\bibitem{EF-DAL-SM:08}
E.~Frew, D.~Lawrence, and S.~Morris, ``Coordinated standoff tracking of moving targets using lyapunov guidance vector fields,'' {\em Journal of Guidance, Control, and Dynamics}, vol.~31, no.~2, pp.~290--306, 2008.

\bibitem{SY-SP-YK:13}
S.~Yoon, S.~Park, and Y.~Kim, ``Circular motion guidance law for coordinated standoff tracking of a moving target,'' {\em Journal of Guidance, Control, and Dynamics}, vol.~36, no.~5, pp.~1454--1465, 2013.

\bibitem{CT-CL-CN:20}
C.~Taylor, C.~Luzzi, and C.~Nowzari, ``On the effects of collision avoidance on emergent swarm behavior,'' in {\em 2020 American Control Conference (ACC)}, pp.~931--936, IEEE, 2020.

\bibitem{OTH:07}
O.~Holland, ``Taxonomy for the modeling and simulation of emergent behavior systems,'' in {\em Proceedings of the 2007 spring simulation multiconference-Volume 2}, pp.~28--35, 2007.

\bibitem{DW:06}
D.~Willshaw, ``Self-organization in the nervous system,'' {\em Cognitive systems: Information processing meet brain science}, pp.~5--33, 2006.

\bibitem{JT:05}
J.~Fromm, ``Types and forms of emergence,'' {\em arXiv preprint nlin/0506028}, 2005.

\bibitem{JT:05ten}
J.~Fromm, ``Ten questions about emergence,'' {\em arXiv preprint nlin/0509049}, 2005.

\bibitem{MG-JC-TJD-RG:14}
M.~Gauci, J.~Chen, T.~J. Dodd, and R.~Gro{\ss}, ``Evolving aggregation behaviors in multi-robot systems with binary sensors,'' in {\em Distributed Autonomous Robotic Systems} (M.~{Ani Hsieh} and G.~Chirikjian, eds.), (Berlin, Heidelberg), pp.~355--367, Springer Berlin Heidelberg, 2014.

\bibitem{DB-RT-OH-SL:18}
D.~S. Brown, R.~Turner, O.~Hennigh, and S.~Loscalzo, ``Discovery and exploration of novel swarm behaviors given limited robot capabilities,'' in {\em Distributed Autonomous Robotic Systems}, Springer International Publishing AG, 2018.

\bibitem{DS-CP-GB:18}
D.~St-Onge, C.~Pinciroli, and G.~Beltrame, ``Circle formation with computation-free robots shows emergent behavioral structure,'' in {\em IEEE/RSJ International Conference on Intelligent Robots and Systems (IROS)}, pp.~5344--5349, IEEE, 2018.

\bibitem{FB-MG-RN:21}
F.~Berlinger, M.~Gauci, and R.~Nagpal, ``Implicit coordination for 3d underwater collective behaviors in a fish-inspired robot swarm,'' {\em Science Robotics}, vol.~6, no.~50, 2021.

\bibitem{RV-KZ-CM-DSB-CN:24}
R.~Vega, K.~Zhu, C.~Mattson, D.~Brown, and C.~Nowzari, ``Agent-based emulation for deploying robot swarm behaviors,'' {\em arXiv e-prints}, p.~arXiv:2410.16444, 2024.

\bibitem{RV-CM-DSB-CN:23}
R.~Vega, C.~Mattson, D.~Brown, and C.~Nowzari, ``Indirect swarm control: Characterization and analysis of emergent swarm behaviors,'' {\em arXiv e-prints}, p.~arXiv:2309.11408, 2023.

\bibitem{HL-ZYH-AH:21}
H.~Litimein, Z.~Huang, and A.~Hamza, ``A survey on techniques in the circular formation of multi-agent systems,'' {\em Electronics}, vol.~10, no.~23, p.~2959, 2021.

\bibitem{AJH-ASMH-DJR-HAA:24}
A.~J. Hepworth, A.~S.~M. Hussein, D.~J. Reid, and H.~A. Abbass, ``Contextually aware intelligent control agents for heterogeneous swarms,'' {\em Swarm Intelligence}, vol.~18, no.~4, pp.~275--310, 2024.

\bibitem{MR-CA-RN:12}
M.~Rubenstein, C.~Ahler, and R.~Nagpal, ``Kilobot: A low cost scalable robot system for collective behaviors,'' in {\em 2012 IEEE international conference on robotics and automation}, pp.~3293--3298, IEEE, 2012.

\bibitem{HH:18}
H.~Hamann, {\em Swarm robotics: A formal approach}, vol.~221.
\newblock Springer, 2018.

\bibitem{AH-AS:22}
A.~Haider and A.~Schmidt, ``Defining the swarm: Challenges in developing nato-agreed terminology across all domains,'' {\em The Journal of the JAPCC}, vol.~34, pp.~60--65, 2022.

\bibitem{RA-KC-BA-CK:19}
R.~Arnold, K.~Carey, B.~Abruzzo, and C.~Korpela, ``What is a robot swarm: a definition for swarming robotics,'' in {\em 2019 IEEE 10th annual ubiquitous computing, electronics \& mobile communication conference (uemcon)}, pp.~0074--0081, IEEE, 2019.

\bibitem{GB-JW:93}
G.~Beni and J.~Wang, ``Swarm intelligence in cellular robotic systems,'' in {\em Robots and biological systems: towards a new bionics?}, pp.~703--712, Springer, 1993.

\bibitem{MS-DGR-MB:24}
M.~Salman, D.~G. Ramos, and M.~Birattari, ``Automatic design of stigmergy-based behaviours for robot swarms,'' {\em Communications Engineering}, vol.~3, no.~1, p.~30, 2024.

\bibitem{MB-AL-DB-MB-GF-LG-DGR-KH-MK-JK:19}
M.~Birattari, A.~Ligot, D.~Bozhinoski, M.~Brambilla, G.~Francesca, L.~Garattoni, D.~G. Ramos, K.~Hasselmann, M.~Kegeleirs, J.~Kuckling, {\em et~al.}, ``Automatic off-line design of robot swarms: a manifesto,'' {\em Frontiers in Robotics and AI}, vol.~6, p.~59, 2019.

\bibitem{ES:04}
E.~{\c{S}}ahin, ``Swarm robotics: From sources of inspiration to domains of application,'' in {\em International workshop on swarm robotics}, pp.~10--20, Springer, 2004.

\bibitem{HA-SM:25}
H.~Abbass and S.~Mostaghim, ``The road forward with swarm systems,'' 2025.

\bibitem{AL-MB:18}
A.~Ligot and M.~Birattari, ``On mimicking the effects of the reality gap with simulation-only experiments,'' in {\em Swarm Intelligence: 11th International Conference, ANTS 2018, Rome, Italy, October 29--31, 2018, Proceedings 11}, pp.~109--122, Springer, 2018.

\bibitem{AK-PW-NC-KS-ML:15}
A.~Kolling, P.~Walker, N.~Chakraborty, K.~Sycara, and M.~Lewis, ``Human interaction with robot swarms: A survey,'' {\em IEEE Transactions on Human-Machine Systems}, vol.~46, no.~1, pp.~9--26, 2015.

\bibitem{MB-EF-MB-MD:13}
M.~Brambilla, E.~Ferrante, M.~Birattari, and M.~Dorigo, ``Swarm robotics: a review from the swarm engineering perspective,'' {\em Swarm Intelligence}, vol.~7, no.~1, pp.~1--41, 2013.

\bibitem{IN-FM:13}
I.~Navarro and F.~Mat{\'\i}a, ``An introduction to swarm robotics,'' {\em International Scholarly Research Notices}, vol.~2013, no.~1, p.~608164, 2013.

\bibitem{CCA-JCBT-RFSO:22}
C.~Calder{\'o}n-Arce, J.~C. Brenes-Torres, and R.~Solis-Ortega, ``Swarm robotics: Simulators, platforms and applications review,'' {\em Computation}, vol.~10, no.~6, p.~80, 2022.

\bibitem{BRJ:10}
B.~R. Johnson, ``Eliminating the mystery from the concept of emergence,'' {\em Biology \& Philosophy}, vol.~25, no.~5, pp.~843--849, 2010.

\bibitem{LSS-PES:78}
L.~Schulman and P.~Seiden, ``Statistical mechanics of a dynamical system based on conway's game of life,'' {\em Journal of Statistical Physics}, vol.~19, pp.~293--314, 1978.

\bibitem{VB:06}
V.~Bauchau, ``Emergence and reductionism: from the game of life to science of life,'' in {\em Self-Organization and Emergence in Life Sciences} (B.~Feltz, M.~Crommelinck, and P.~Goujon, eds.), pp.~29--40, Springer, 2006.

\bibitem{SW:02}
S.~Wolfram, {\em A New Kind of Science}.
\newblock Champaign, IL: Wolfram Media, 2002.

\bibitem{SC-AAP-PD-SS-VK:18}
S.~Chung, A.~A. Paranjape, P.~Dames, S.~Shen, and V.~Kumar, ``A survey on aerial swarm robotics,'' {\em IEEE Transactions on Robotics}, vol.~34, no.~4, pp.~837--855, 2018.

\bibitem{LY-JY-SY-BW-BJN-LZ:21}
L.~Yang, J.~Yu, S.~Yang, B.~Wang, B.~J. Nelson, and L.~Zhang, ``A survey on swarm microrobotics,'' {\em IEEE Transactions on Robotics}, vol.~38, no.~3, pp.~1531--1551, 2021.

\bibitem{LI-DN-MS:01}
L.~Iocchi, D.~Nardi, and M.~Salerno, ``Reactivity and deliberation: A survey on multi-robot systems,'' in {\em Balancing Reactivity and Social Deliberation in Multi-Agent Systems}, (Berlin, Heidelberg), pp.~9--32, Springer Berlin Heidelberg, 2001.

\bibitem{MS-MU-MS-WE:20}
M.~Schranz, M.~Umlauft, M.~Sende, and W.~Elmenreich, ``Swarm robotic behaviors and current applications,'' {\em Frontiers in Robotics and AI}, vol.~7, p.~36, 2020.

\bibitem{MD-GT-VT:21}
M.~Dorigo, G.~Theraulazand, and V.~Trianni, ``Swarm robotics: past, present, and future [point of view],'' {\em Proceedings of the IEEE}, vol.~109, no.~7, pp.~1152--1165, 2021.

\bibitem{ARC-SS-KG:22}
A.~R. Cheraghi, S.~Shahzad, and K.~Graffi, ``Past, present, and future of swarm robotics,'' in {\em Intelligent Systems and Applications: Proceedings of the 2021 Intelligent Systems Conference (IntelliSys) Volume 3}, pp.~190--233, Springer, 2022.

\bibitem{YM-SGP:09}
Y.~Mohan and S.~G. Ponnambalam, ``An extensive review of research in swarm robotics,'' in {\em 2009 world congress on nature \& biologically inspired computing (nabic)}, pp.~140--145, IEEE, 2009.

\bibitem{JCB-YAK:13}
J.~C. Barca and Y.~A. Sekercioglu, ``Swarm robotics reviewed,'' {\em Robotica}, vol.~31, no.~3, pp.~345--359, 2013.

\bibitem{MMS-ZS-AA-HM-MHY-NKB-FH:23}
M.~M. Shahzad, Z.~Saeed, A.~Akhtar, H.~Munawar, M.~H. Yousaf, N.~K. Baloach, and F.~Hussain, ``A review of swarm robotics in a nutshell,'' {\em Drones}, vol.~7, no.~4, p.~269, 2023.

\bibitem{RD-HS-OM:13}
R.~Doursat, H.~Sayama, and O.~Michel, ``A review of morphogenetic engineering,'' {\em Natural Computing}, vol.~12, no.~4, pp.~517--535, 2013.

\bibitem{YJ-YM:11}
Y.~Jin and Y.~Meng, ``Morphogenetic robotics: An emerging new field in developmental robotics,'' {\em IEEE Transactions on Systems, Man, and Cybernetics, Part C: Applications and Reviews}, vol.~41, no.~2, pp.~145--160, 2011.

\bibitem{RD-HS-OM:12}
R.~Doursat, H.~Sayama, and O.~Michel, {\em Morphogenetic engineering: toward programmable complex systems}.
\newblock Springer, 2012.

\bibitem{TJ:08}
T.~Jansen, ``Strandbeests,'' {\em Architectural Design}, vol.~78, no.~4, pp.~22--27, 2008.

\bibitem{SP-15}
S.~Patnaik, ``Analysis of theo jansen mechanism (strandbeest) and its comparative advantages over wheel based mine escavation system,'' {\em IOSR J Eng}, vol.~5, no.~7, pp.~43--52, 2015.

\bibitem{JH-MB:21}
J.~D. Hasbach and M.~Bennewitz, ``The design of self-organizing human–swarm intelligence,'' {\em Adaptive Behavior}, vol.~30, no.~4, pp.~361--386, 2021.

\bibitem{HS:09}
H.~Sayama, ``Swarm chemistry,'' {\em Artificial life}, vol.~15, no.~1, pp.~105--114, 2009.

\bibitem{HH-KJ-JL:21}
H.~Haeri, K.~Jerath, and J.~Leachman, ``Thermodynamics-inspired macroscopic states of bounded swarms,'' {\em ASME Letters in Dynamic Systems and Control}, vol.~1, no.~1, p.~011015, 2021.

\bibitem{CM-DB:23}
C.~Mattson and D.~Brown, ``Leveraging human feedback to evolve and discover novel emergent behaviors in robot swarms,'' in {\em Proceedings of the Genetic and Evolutionary Computation Conference}, pp.~56--64, 2023.

\bibitem{CM-VR-RV-CN-DSD-DSB:25}
C.~Mattson, V.~Raveendra, R.~Vega, C.~Nowzari, D.~S. Drew, and D.~S. Brown, ``Discovery and deployment of emergent robot swarm behaviors via representation learning and real2sim2real transfer,'' {\em arXiv preprint arXiv:2502.15937}, 2025.

\bibitem{JG-ALC:18}
J.~Gomes and A.~Christensen, ``Task-agnostic evolution of diverse repertoires of swarm behaviours,'' in {\em Swarm Intelligence: 11th International Conference, ANTS 2018, Rome, Italy, October 29--31, 2018, Proceedings 11}, pp.~225--238, Springer, 2018.

\bibitem{JG-PU-ALC:13}
J.~Gomes, P.~Urbano, and A.~L. Christensen, ``Evolution of swarm robotics systems with novelty search,'' {\em Swarm Intelligence}, vol.~7, no.~2, pp.~115--144, 2013.

\bibitem{FC-ZC-LX-ZXL-ZY:09}
F.~Chen, Z.~Chen, L.~Xiang, Z.-X. Liu, and Z.~Yuan, ``Reaching a consensus via pinning control,'' {\em Automatica}, vol.~45, no.~5, pp.~1215--1220, 2009.

\bibitem{XFW-GC:02}
X.~Wang and G.~Chen, ``Pinning control of scale-free dynamical networks,'' {\em Physica A: Statistical Mechanics and its Applications}, vol.~310, no.~3–4, p.~521–531, 2002.

\bibitem{XW-XL-JL:24}
X.~Wang, X.~Li, and J.~Lu, ``Control and flocking of networked systems via pinning,'' {\em IEEE Circuits and Systems Magazine}, vol.~10, no.~3, pp.~83--91, 2024.

\bibitem{JW-LM:07}
J.~Wang and L.~M. Lewis, ``Assessing coordination overhead in control of robot teams,'' in {\em Proceedings of the IEEE International Conference on Systems, Man and Cybernetics}, pp.~2645--2650, IEEE, 2007.

\bibitem{AH-HAA:14}
A.~Hussein and H.~Abbass, ``Human-swarm interaction: sources of uncertainty,'' in {\em Proceedings of the 2014 ACM/IEEE International Conference on Human-Robot Interaction}, pp.~166--167, ACM, 2014.

\bibitem{PW-SN-ML-AK-NC-KS:12}
P.~Walker, S.~Nunnally, M.~Lewis, A.~Kolling, N.~Chakraborty, and K.~Sycara, ``Neglect benevolence in human control of swarms in the presence of latency,'' in {\em Proceedings of IEEE International Conference on Systems, Man, and Cybernetics (SMC '12)}, pp.~309--314, IEEE, 2012.

\bibitem{LP-PB-WPMHM:25}
L.~Pedroso, P.~Batista, and W.~P. M. H.~M. Heemels, ``Distributed design of ultra large-scale control systems: Progress, challenges, and prospects,'' {\em Annual Reviews in Control}, vol.~59, p.~100987, 2025.

\bibitem{YT-ZZ:13}
Y.~Tan and Z.~Zheng, ``Research advance in swarm robotics,'' {\em Defence Technology}, vol.~9, no.~1, pp.~18--39, 2013.

\bibitem{BK-FC:15}
B.~Khaldi and F.~Cherif, ``An overview of swarm robotics: Swarm intelligence applied to multi-robotics,'' {\em International Journal of Computer Applications}, vol.~126, no.~2, 2015.

\bibitem{AMS-MK:16}
A.~M. Schroeder and M.~Kumar, ``Design of decentralized chemotactic control law for area coverage using swarm of mobile robots,'' in {\em 2016 American Control Conference (ACC)}, pp.~4317--4322, IEEE, 2016.

\bibitem{AD-MK-SR:17}
A.~Deshpande, M.~Kumar, and S.~Ramakrishnan, ``Robot swarm for efficient area coverage inspired by ant foraging: the case of adaptive switching between brownian motion and l{\'e}vy flight,'' in {\em Dynamic Systems and Control Conference}, vol.~58288, p.~V002T14A009, American Society of Mechanical Engineers, 2017.

\bibitem{HJML-DAL:22}
H.~J.~M. Lopes and D.~A. Lima, ``Surveillance task optimized by evolutionary shared tabu inverted ant cellular automata model for swarm robotics navigation control,'' {\em Results in Control and Optimization}, vol.~8, p.~100141, 2022.

\bibitem{AO-MG-AK-MDH-RG:19}
A.~{\"O}zdemir, M.~Gauci, A.~Kolling, M.~Hall, and R.~Gro{\ss}, ``Spatial coverage without computation,'' in {\em 2019 International Conference on Robotics and Automation (ICRA)}, pp.~9674--9680, IEEE, 2019.

\bibitem{QW-HZ:21}
Q.~Wang and H.~Zhang, ``A self-organizing area coverage method for swarm robots based on gradient and grouping,'' {\em Symmetry}, vol.~13, no.~4, p.~680, 2021.

\bibitem{DAL-GMBO:17}
D.~A. Lima and G.~M.~B. Oliveira, ``A cellular automata ant memory model of foraging in a swarm of robots,'' {\em Applied Mathematical Modelling}, vol.~47, pp.~551--572, 2017.

\bibitem{RZ-VK-EW-JP-SB:15}
R.~Zou, V.~Kalivarapu, E.~Winer, J.~Oliver, and S.~Bhattacharya, ``Particle swarm optimization-based source seeking,'' {\em IEEE Transactions on Automation Science and Engineering}, vol.~12, no.~3, pp.~865--875, 2015.

\bibitem{ASK-GM-RRB-MSC:17}
A.~S. Kumar, G.~Manikutty, R.~R. Bhavani, and M.~S. Couceiro, ``Search and rescue operations using robotic darwinian particle swarm optimization,'' in {\em 2017 International Conference on Advances in Computing, Communications and Informatics (ICACCI)}, pp.~1839--1843, IEEE, 2017.

\bibitem{RM-CSJ-CV:18}
R.~Melo, C.~S. Junior, and C.~Victor, ``An performance analysis of a pso-based algorithm for swarm robotics,'' in {\em Proc. 9th Int. Multi-Conf. Complex., Inform. Cybern.(IMCIC)}, pp.~1--7, 2018.

\bibitem{JY-RX-XX-YS:20}
J.~Yang, R.~Xiong, X.~Xiang, and Y.~Shi, ``Exploration enhanced rpso for collaborative multitarget searching of robotic swarms,'' {\em Complexity}, vol.~2020, no.~1, p.~8863526, 2020.

\bibitem{VG-AS-RT:22}
V.~Garg, A.~Shukla, and R.~Tiwari, ``Aerpso—an adaptive exploration robotic pso based cooperative algorithm for multiple target searching,'' {\em Expert Systems with Applications}, vol.~209, p.~118245, 2022.

\bibitem{BY-YD-YJ-KH:15}
B.~Yang, Y.~Ding, Y.~Jin, and K.~Hao, ``Self-organized swarm robot for target search and trapping inspired by bacterial chemotaxis,'' {\em Robotics and Autonomous Systems}, vol.~72, pp.~83--92, 2015.

\bibitem{OYM:21}
O.~Y. Maryasin, ``Bee-inspired algorithm for groups of cyber-physical robotic cleaners with swarm intelligence,'' {\em Cyber-Physical Systems: Modelling and Intelligent Control}, pp.~167--177, 2021.

\bibitem{ZZ-YT:13}
Z.~Zheng and Y.~Tan, ``Group explosion strategy for searching multiple targets using swarm robotic,'' in {\em 2013 IEEE Congress on evolutionary computation}, pp.~821--828, IEEE, 2013.

\bibitem{JPM-CN:23}
J.~P. Mathew and C.~Nowzari, ``Real-time distributed infrastructure-free searching and target tracking via virtual pheromones,'' {\em arXiv preprint arXiv:2311.13035}, 2023.

\bibitem{JL-YT:14}
J.~Li and Y.~Tan, ``The multi-target search problem with environmental restrictions in swarm robotics,'' in {\em 2014 IEEE International Conference on Robotics and Biomimetics (ROBIO 2014)}, pp.~2685--2690, IEEE, 2014.

\bibitem{ZZ-JL-JL-YT:14}
Z.~Zheng, J.~Li, J.~Li, and Y.~Tan, ``Avoiding decoys in multiple targets searching problems using swarm robotics,'' in {\em 2014 IEEE Congress on Evolutionary Computation (CEC)}, pp.~784--791, IEEE, 2014.

\bibitem{QT-FY-LD:16}
Q.~Tang, F.~Yu, and L.~Ding, ``A grouping method for multiple targets search using swarm robots,'' in {\em Advances in Swarm Intelligence: 7th International Conference, ICSI 2016, Bali, Indonesia, June 25-30, 2016, Proceedings, Part II 7}, pp.~470--478, Springer, 2016.

\bibitem{JL-YT:16}
J.~Li and Y.~Tan, ``Triangle formation based multiple targets search using a swarm of robots,'' in {\em International Conference in Swarm Intelligence}, pp.~544--552, Springer, 2016.

\bibitem{ZN-QZ-XW-FW-TH:23}
Z.~Nie, Q.~Zhang, X.~Wang, F.~Wang, and T.~Hu, ``Triangular lattice formation in robot swarms with minimal local sensing,'' {\em IET Cyber-Systems and Robotics}, vol.~5, no.~2, p.~e12087, 2023.

\bibitem{MG-JC-WL-TJD-RG:14}
M.~Gauci, J.~Chen, W.~Li, T.~Dodd, and R.~Gro{\ss}, ``Self-organized aggregation without computation,'' {\em The International Journal of Robotics Research}, vol.~33, no.~8, pp.~1145--1161, 2014.

\bibitem{VG:05}
V.~Gazi, ``Swarm aggregations using artificial potentials and sliding-mode control,'' {\em IEEE Transactions on Robotics}, vol.~21, no.~6, pp.~1208--1214, 2005.

\bibitem{DS-LV:19}
D.~Shah and L.~Vachhani, ``Swarm aggregation without communication and global positioning,'' {\em IEEE Robotics and Automation Letters}, vol.~4, no.~2, pp.~886--893, 2019.

\bibitem{JW-CM:03}
J.~Wessnitzer and C.~Melhuish, ``Collective decision-making and behaviour transitions in distributed ad hoc wireless networks of mobile robots: Target-hunting,'' in {\em European Conference on Artificial Life}, pp.~893--902, Springer, 2003.

\bibitem{SG-JG-MA-CJ-GT:09}
S.~Garnier, J.~Gautrais, M.~Asadpour, C.~Jost, and G.~Theraulaz, ``Self-organized aggregation triggers collective decision making in a group of cockroach-like robots,'' {\em Adaptive Behavior}, vol.~17, no.~2, pp.~109--133, 2009.

\bibitem{AC-SG-OD-MZ-MD:11}
A.~Campo, S.~Garnier, O.~D{\'e}driche, M.~Zekkri, and M.~Dorigo, ``Self-organized discrimination of resources,'' {\em PLoS One}, vol.~6, no.~5, p.~e19888, 2011.

\bibitem{AG-AC-FM-LM-MD:10}
A.~Guti{\'e}rrez, A.~Campo, F.~Monasterio-Huelin, L.~Magdalena, and M.~Dorigo, ``Collective decision-making based on social odometry,'' {\em Neural Computing and Applications}, vol.~19, pp.~807--823, 2010.

\bibitem{CACP-HZ:11}
C.~A.~C. Parker and H.~Zhang, ``Biologically inspired collective comparisons by robotic swarms,'' {\em The International Journal of Robotics Research}, vol.~30, no.~5, pp.~524--535, 2011.

\bibitem{AA-MA-JC-ET:21}
A.~Almansoori, M.~Alkilabi, J.~Colin, and E.~Tuci, ``On the evolution of mechanisms for collective decision making in a swarm of robots,'' in {\em Italian Workshop on Artificial Life and Evolutionary Computation}, pp.~109--120, Springer, 2021.

\end{thebibliography}

\end{document}